\documentclass[pdflatex,default,iicol]{sn-jnl}% Default with double column layout

\usepackage{graphicx}%
\usepackage{multirow}%
\usepackage{amsmath,amssymb,amsfonts}%
\usepackage{amsthm}%
\usepackage{mathrsfs}%
\usepackage[title]{appendix}%
\usepackage{xcolor}%
\usepackage{textcomp}%
\usepackage{manyfoot}%
\usepackage{booktabs}%
\usepackage{algorithm}%
\usepackage{algorithmicx}%
\usepackage{algpseudocode}%
\usepackage{listings}%
\usepackage{caption}
\usepackage{subcaption}
\usepackage{mwe}
\usepackage{placeins}
\usepackage{booktabs}
\usepackage{pifont}
\usepackage{xspace}
\usepackage[T1]{fontenc}%for references

\usepackage{threeparttable}
\usepackage{rotating}
\usepackage{multirow}
\usepackage{tabularx}
\usepackage{amsmath}
\usepackage{ gensymb }
\usepackage{ wasysym }
\usepackage[nolist]{acronym}
\usepackage{calc}

\theoremstyle{thmstyleone}%
\theoremstyle{thmstyletwo}%

\theoremstyle{thmstylethree}%

\begin{document}

\title[Automated Deception Detection from Videos]{Automated Deception Detection from Videos: Using End-to-End Learning Based High-Level Features and Classification Approaches}


\author*[1]{\fnm{Laslo} \sur{Dinges}}\email{laslo.dinges@ovgu.de} 
\author[1]{\fnm{Marc-André} \sur{Fiedler}}\email{\{firstname.lastname\}@ovgu.de}
\author[1]{\fnm{Ayoub} \sur{Al-Hamadi}}
\author[1]{\fnm{Thorsten} \sur{Hempel}}
\author[1]{\fnm{Ahmed} \sur{Abdelrahman}}
\author[2]{\fnm{Joachim} \sur{Weimann}}
\author[2]{\fnm{Dmitri} \sur{Bershadskyy}}

%\author[2,3]{\fnm{Second} \sur{Author}}


\affil*[1]{\orgdiv{Neuro-Information Technology Group}, \orgname{Otto-von-Guericke University}, \orgaddress{\street{P.O. Box 4210}, \city{Magdeburg}, \postcode{D-39106}, \state{Saxony-Anhalt}, \country{Germany}}}

\affil[2]{\orgdiv{Faculty of Economics and Management}, \orgname{Otto-von-Guericke University}, \orgaddress{\street{P.O. Box 4210}, \city{Magdeburg}, \postcode{D-39106}, \state{Saxony-Anhalt}, \country{Germany}}}


\abstract{
Deception detection is an interdisciplinary field attracting researchers from psychology, criminology, computer science, and economics. Automated deception detection presents unique challenges compared to traditional polygraph tests, but also offers novel economic applications. In this spirit, we propose a multimodal approach combining deep learning with discriminative models for deception detection.
 Video modalities are analyzed using convolutional end-to-end learning for gaze, head pose, and facial expressions, achieving promising results compared to state-of-the-art approaches.
Due to the limited availability of training data, rendering end-to-end learning scarcely feasible, we then employ discriminative models to detect deception. Sequence-to-Class approaches are also explored but outperformed by discriminative models due to data scarcity.
We evaluate our approach on five datasets, including four well-known publicly available datasets and a new economically motivated Rolling-Dice Experiment.
Results reveal performance differences among modalities, with facial expressions outperforming gaze and head pose overall. Combining multiple modalities and feature selection consistently enhances detection performance.
The observed variations in expressed features across datasets with different contexts affirm the importance of scenario-specific training data for effective deception detection, further indicating the influence of context on deceptive behavior. Cross-dataset experiments reinforce these findings.
Notably, low-stake datasets, including the Rolling-Dice Experiment, present more challenges for deception detection compared to the high-stake Real-Life trials dataset. Nevertheless, various evaluation measures show deception detection performance surpassing chance levels.
Our proposed multimodal approach and comprehensive evaluation highlight the challenges and potential of automating deception detection from video modalities, offering promise for future research.
}

\keywords{Deception detection, Multimodal, End-to-end, Rolling-dice}



\maketitle

%%%%%%%%%%%%%%%%%%%%%%%%%%%%%%%%%%%%%%%%%%%%%%%%
%%Variables & Makros
\newlength{\subfigW}
\setlength{\subfigW}{8.6cm}

%%makros
\newcommand{\cmark}{\ding{51}}%
\newcommand{\xmark}{\ding{55}}%

%pseudo text as placeholder
\newif\ifuselorum % Define a new boolean variable
\uselorumfalse % Set the boolean variable to true

%show tables with intermediate results
\newif\ifextratables % Define a new boolean variable
\extratablesfalse % Set the boolean variable to true

\definecolor{myGreen}{RGB}{118,183,39} 

% Measure the total width of the text area
\newlength{\totalwidth}
\setlength{\totalwidth}{\textwidth}
% Calculate the width of a single column
\newlength{\singlecolumnwidth}
\setlength{\singlecolumnwidth}{(\totalwidth-\columnsep)/2}

%TODO: add acronyms 
\begin{acronym}
        \acro{dice}[RDE]{Rolling-dice Experiment}
        \acro{bag}[BgL]{Bag-of-lies}
        \acro{box}[BxL]{Box-of-lies}
        \acro{rl}[RL]{Real-life trial}
        \acro{mu}[MU3D]{Miami University Deception Detection Database}       
        \acro{me}[ME]{micro-expressions}
	\acro{gt}[GT]{ground truth}
	\acro{va}[VA]{valence \& arousal}
	\acro{au}[AU]{Action-Unit}
        \acro{ml}[ML]{Machine-Learning}        
	\acro{svr}[SVR]{Support-Vector-Regression}
	\acro{svm}[SVM]{Support-Vector-Machine}
        \acro{rf}[RF]{Random-Forest}
        \acro{cnn}[CNN]{Convolutional-Neural-Network}
	\acro{circ}[V/A-circ. m.]{Valence\&Arousal-circumplex model}
	\acro{lstm}[LSTM]{Long-Short-Term-Memory network}	
        \acro{rnn}[RNN]{Recurrent-Neural-Network}
	\acro{hrc}[HRC]{Human-Robot Collaboration}
        \acro{roc}[RoC]{Receiver Operating Characteristic }
        \acro{auc}[AUC]{Area-Under-Curve}
\end{acronym}
%%%%%%%%%%%%%%%%%%%%%%%%%%%%%%%%%%%%%%%%%%%%%%%%

%textwidth = 6.3 inch = 160mm  , single column = 3inch = 76mm

\section{Introduction}\label{sec1}

The study of deception detection is a complex and multidisciplinary field that has gathered significant research attention. Several kinds of techniques have been proposed to detect signs of deception, including traditional methods such as polygraph tests and modern approaches utilizing computer vision, natural language processing, and machine learning~\cite{alaskar2022intelligent}. Although the primary applications of deception detection techniques have traditionally been in the field of criminalistics, with a focus on interrogation, recent developments have shown that it has also become a significant concern in other domains such as border security~\cite{sanchez2022politics}, where automized contact-free or even online approaches are preferred over traditional polygraph tests, which are too complicated and costly for mass screening. 
However, using detection approaches in such a way -- that might lead to conviction or the restriction of the freedom to travel -- is delicate due to possible unreliable or discriminating decisions.

%economy
The significance of digitization and machine learning-based approaches also has now become a critical issue in the field of experimental economics~\cite{brynjolfsson2021economics,weimann2019methods,camerer2019replication}.
In this regard, also automized deception detection is of interest and, in contrast to the aforementioned domains, it can be used in a less invasive way.
%% the problem, damage to society (extreme cases)
%Product piracy serves as an extreme example of the economic damage caused by deception since such products often lack even the most basic safety requirements. Other examples are deceit in the industry, politics, or even medicine.
%our specific case
Salespeople who misrepresent the quality of their products can cause significant economic harm. Such fraudulent activities not only affect the buyers but also harm salespeople who offer high-quality products but suffer losses due to customers' cautious purchasing behaviors as a result of bad experiences~\cite{butski2022honest}. 
People often overestimate their ability to detect lies, when in reality their accuracy is only slightly better than chance, and at the same time they underestimate their ability to tell lies.
Hence, an automated deception detection system could be useful, for example, to assist customers
 during virtual sales meetings using video and audio signals from regular webcams.
Although accuracy may be lower compared to professional polygraph tests (which are hardly feasible in this context), they can still provide a valuable contribution to society by warning customers of potential deception as greatly exaggerated product quality.

In the following, we focus on the lie -- that can be a wrong description of a photo or object that was shown to subjects or a false statement in a legal context -- while other forms of deception as concealment, evasion, gaslighting, or trick scam are beyond the scope of this work. 
However, because the used databases include more than binary responses, they may also include half-truths, misleading, exaggeration, or manipulation. Furthermore, also the investigated signs are not restricted to a single modality, since many features such as facial expression, voice, transcription, vital parameters, and body language might be cues for deceptions as lies.
As a matter of fact, we propose a multi-modal approach 
%that is based on several deep-learning models 
to evaluate some of the most promising features of contact-free deception detection on the four publicly available deception databases. 
Afterward, we employ this approach to evaluate a classical experiment known as the dice-rolling experiment~\cite{fischbacher2013lies}.

\section{Related Works}
In the following, we give an overview of related work in the area of general deception detection as well as from the point of view of AI-based automation.

\subsection{Polygraph Tests}
%todo Lie detection (in general, psychological papers or polygraph applications)
Polygraph testing is one of the most widely used techniques for detecting deception. 
Polygraph tests measure physiological responses such as heart rate, blood pressure, and skin conductivity to determine whether a person is lying. Despite their widespread use, the accuracy of polygraph tests has been a topic of debate in the scientific community~\cite{nortje2019good}.

There are several types of polygraph tests, such as the Control Question Technique (CQT) or the less widely used Concealed Information Test (CIT). 
These, however, do not differ in the kind of features that are used to detect deception, but in the kind of questions that are asked during interrogation. Unlike CIT, the CQT can be used even without critical information, however, CQT has been even more criticized than CIT due to its unethical testing conditions and highly questionable assumptions~\cite{nortje2019good}.

In general, polygraphs tests are judged clearly less suitable by the scientific community than by the American Polygraph Association, for example since  deceptive answers will not necessarily produce unique cues or might be sensitive to cultural or other context~\cite{herbig2020psychology,sanchez2022politics}. 
% CQT decept rec. 84%–89%  truth rec. 59%–75%
%
Another issue is that truth-tellers more often believe that their innocence is obvious (known as the illusion of transparency), which can make truth-tellers even less credible, for example since they tend to react more aggressively than deceptive people if they feel they are not believed~\cite{zloteanu2020reconsidering}. 
This is one of many reasons why traditional polygraph tests -- but also automized deception detection -- will only indicate the possibility of deception in real-life scenarios, rather than providing a definitive identification. 
However, these methods are still clearly better than chance, while purely manual attempts are not~\cite{nortje2019good,gupta2019bag}.

\subsection{Automized Deception Detection}
In the following, we review some key works of contact-free, automized uni- and multi-modal deception detection. 

%Automized / Multi-Modal (main:Video)
While polygraph tests are criticized but still relatively reliable due to contact measurements such as skin conductance and, even more important, due to an expert that interrogates the candidates, approaches of automized deception detection are typically limited to contact-free video or audio data modalities and available databases do not involve any baselines as control questions.
Since available databases are also not comprehensive, automized detection of deception is even more challenging.
However, since they are simple to use and involve much lower cost, also new areas of applications are enabled, for example as online service.
 %Overeview of modalitier
 Typical contact-free modalities which are used for automized deception detection are micro-expressions, macro-expressions, thermal images, gaze, gestures, voice features (tone and pitch), or the transcription of what was said~\cite{saini2022ldm}.

%%  END-TO-END for feature generation, some paper state all typical features
%Emotion and Action Units

\subsubsection{Gaze and Headpose}
%Now: describe SOTA of all or most important features
%Headpose 
% why no body pose
While body poses and gestures can be effective features for detecting deception in certain contexts~\cite{avola2020lietome}, it may not be the case in all situations. For instance, in our targeted application scenario and most of the available deception databases, subjects are  seated and do not exhibit significant changes in their body language or hand-gestures that can be used for deception detection purposes.
However, head pose and gaze direction can also serve as potential indicators of the user's state~\cite{schepisi2020oculomotor}.

%Headpose - Applications
\paragraph{Headpose}
Head pose estimation from a single image is a crucial task for various applications, including driver assistance, human-robot interaction, and even pain detection~\cite{werner2013towards}. 
Apart from that, people often avoid eye contact when they feel guilty (for example after being untruthful), which makes tracking changes in head pose and gaze direction valuable for lie detection purposes~\cite{schepisi2020oculomotor}.
%types
Two types of methods are commonly used for this task: landmark-based and landmark-free. Landmark-based methods require accurate landmark detection for accurate 3D head pose estimation~\cite{8297015}, while landmark-free methods estimate head pose directly using deep neural networks to formulate orientation prediction as an appearance-based task~\cite{Ruiz2018FineGrainedHP,8444061,Huang2020ImprovingHP}. 
 Landmark-based methods can be affected by occlusion and extreme rotation, which is why landmark-free approaches perform better on comprehensive but challenging datasets.
 %There are various approaches for this, such as multi-loss, cross-entropy, and feature decoupling networks.

%Gaze
\paragraph{Gaze}
Gaze estimation can be achieved by conventional regression-based methods; however, recently, approaches which are based on \acp{cnn} are favored. 
\acp{cnn} have shown promise in modeling the nonlinear mapping function between images and gaze, with researchers proposing different architectures that take into account factors such as eye images, full-face images, and head pose~\cite{zhang2017mpiigaze}.
Other approaches include combining statistical models with deep learning and using temporal models such as \acp{lstm}~\cite{xiong2019mixed,kellnhofer2019gaze360}.
Another work is AGE-Net, which uses two parallel networks for each eye image and an attention-based network to generate a weight feature vector, with outputs multiplied and refined with the output of VGG as CNN of the face images~\cite{biswas2021appearance}. 

 In the context of deception detection, several studies have utilized gaze features.
Kumar et al. extracted statistical features from the gaze of a group of individuals playing the game Resistance~\cite{kumar2021deception}.
 They found that deceivers change their focus less often than non-deceivers, indicating less engagement in the game.
%pupil delation might be usefull
Beyond gaze direction, Pasquali et al. report that the mean pupil dilation is a good indication of deception of children when playing a game. Considering that only a single feature was used, they received the promising lie-classification F1-Score of 56.5\% for \ac{rf}, respectively 67.7\%  for \ac{svm}~\cite{pasquali2021detecting}.
However, they did not compare this feature with any others, such as gaze direction or blinking. 
Furthermore, their approach requires additional, not contact-free hardware (Tobo Pro Glasses eye-tracker) to ensure accurate measurement of pupil dilation. 
Gupta et al. use the PyGaze library to extract fixations, eye blinks and
pupil size as features. They got 57.11\% accuracy on the bag-of-lies database which is better than accuracy for video (Local Binary Pattern), audio (frequency features) and EEG modality.
 Combining features from all modalities, 66\% accuracy were achieved and 60\% using all contact-free modalities~\cite{gupta2019bag}.

\subsubsection{Facial Expressions}
%Emotions
Facial expressions are a common way to detect basic emotions, which are typically conveyed through specific sets of facial movements known as Action Units (AUs)~\cite{zloteanu2020reconsidering}. These AUs may be either macro- or micro-expressions. 

%Micro Expressions
\paragraph{Micro-expressions}
Micro-expressions are brief, involuntary facial expressions that occur when a person tries to conceal their true feelings or intentions~\cite{ben2021video}. They last less than half a second, making them difficult to detect. However, thanks to technological advances, it is now possible to automatically detect and analyze micro-expressions, so they could be used for lie detection \cite{wu2018deception}.

Some studies have shown that (manually detected) micro-expressions (ME) can be used for accurate detection of deception, as they reveal emotions which a person tries to hide ~\cite{frank2015microexpressions,choi2018}.
%choi: LMF = relative distances between landmark features -> VGG16 -> LSTM 
However, it should be noted that micro-expressions do not always indicate deception, as they can also be caused by other factors such as stress or anxiety. Therefore, automatic deception detection systems that rely on micro-expressions should be used in conjunction with other methods to achieve the most accurate results~\cite{zloteanu2020reconsidering}.
%-> so this is why ME might not work for lie detection
According to Jordan et al., the practical use of ME for lie detection is even questionable, and it may not enhance the overall accuracy of the results~\cite{jordan2019test}. 
One reason why using micro-expressions (ME) for lie detection is limited is that they occur in both genuine and deceptive reactions, 
and occur far too infrequently and in ways that are too variable to be useful in detecting deception \cite{jordan2019test,zloteanu2020reconsidering}. 
Also, the infrequent occurrence reduces its quality as a feature for manual, computer-assisted or automized lie detection. 
Hence, even with a model for ME detection or classification that generalizes well -- so it could be used beyond controlled laboratory conditions and with 30-60 instead of 100-200 frames-per-second(fps)-- applicability of ME in real-life deception detection scenarios would be still limited at the present time.
Additionally, even training such a model for detecting and classifying ME within unseen data in the first place is challenging:
The available databases (as CASME II or SAMM) are limited in sense of comprehensiveness, general applicable experimental design, and kind of covered Action-Units ~\cite{talluri2022,zhang2021review}. %add ref for detection (tallluri covers only classif.)

%Makro-Expressions, general
\paragraph{Macro-expressions}
Macro-expressions are regular facial expressions described by the facial Action-Unit (AU) coding system~\cite{zhi2020comprehensive}. 
AUs as Cheeck Raiser (AU6) or Brow Lowerer (AU4) measure observable muscle activations in specific regions of the face and can be used to predict underlying basic emotions, pain or other secondary emotions from individual or sequences of images.
In-the-wild, macro-expressions can sometimes result in AU of weak intensity, but will still last longer than micro-expressions.
%Circumplex
Since the reliable, unambiguous assignment to exactly one basic emotion or neutral is barely feasible for many in-the-wild samples, the regression of continuous values for valence and arousal values is used instead or additionally, following the idea of the circumplex model~\cite{russell1980circumplex}.

%Emotion recognition
Previous methods for facial expression recognition mixing deep learning with other methods, following a conventional pipeline of face recognition, landmark extraction, and  regression of AUs, which are then interpreted by a classifier to derive expressed basic or secondary emotions as pain~\cite{werner2017facial,vinkemeier2018predicting}.
In this regard, training several partial solutions on -- in some cases small and under controlled conditions acquired -- databases is problematic, as for example, landmark extraction might fail when applied to challenging in-the-wild samples.
Hence, similar to head pose detection, landmark-free approaches that estimate AUs~\cite{werner2020facial}, basic emotions or valence/arousal values directly from raw images (end-to-end learning) are more effective on challenging datasets, provided there are enough available training samples~\cite{handrich2020simultaneous}.

%examples
In their study, Chang et al.~\cite{chang2017fatauva} used a \ac{cnn} to compute AUs, which were subsequently used to predict valence and arousal values. 
On the other hand, Khorrami et al.~\cite{khorrami2015deep} used a holistic strategy to directly predict valence and arousal values from (normalized) face images. AUs were implicitly extracted using a \ac{cnn}.

Mollahosseini et al. proposed the comprehensive AffectNet database and compared traditional approaches as HoG features and \ac{svr} resp. \ac{svm}  with an end-to-end learning approach on it, where end-to-end clearly outperforms the other approaches~\cite{mollahosseini2017affectnet}.
Zhang et al. took a similar approach, where they used a pre-trained network to make predictions about emotions and personalities~\cite{zhang2019persemon}. Li et al. trained a bidirectional RNN that incorporated future images for prediction~\cite{li2017estimation}. 

In the study of Chu et al., they proposed to extract both spatial representations from the data using a CNN and temporal representations using a \ac{lstm}~\cite{chu2017learning}. Their approach merged the results of the \ac{cnn} and \ac{lstm} models, resulting in better performance in predicting AUs. However, no further emotion classification was performed in their study.

\subsubsection{Other Modalities}
A special kind of feature can be extracted from thermal cameras.  
Satpathi et al.~\cite{satpathi2020detection} derive the normalized blood flow rate from regions of interest (forehead and periorbital region) of such thermal data from 10 subjects. 
They found, that blood flow raising steeper for subjects which lie.
However, as expensive, specialized equipment, thermal cameras are barely suitable for online applications as virtual sales meetings.
Hence, it is of essential interest if such features can also be predicted using plain RGB-cameras.
This would still imply some requirements -- as good video quality and low variations of light, face pose, etc. -- but is indeed possible for vital parameters as heart or respiratory rate~\cite{fiedler2021facial}.

%TRanscriptions
When it comes to transcriptions, it's worth noting that individuals who engage in deception tend to be vaguer, containing less self-references and more negative and cognitive
mechanism words~\cite{ioannidis2020lie,bond2005language}.
While providing specific details can enhance credibility, it also increases the risk of being disproved. This approach may be effective in a legal context or similar situations, however, it might be less  useful if a deceiver just has to describe for example a product in one or two sentences without any interrogation.
Ioannidis et al. give a detailed mathematical description of a statistical model  for transcription-based lie detection for legal context \cite{ioannidis2020lie}.
However, this work is primarily theoretical and does not provide any concrete results that can be compared with other methods on a specific database.

%IBorder
IBorderCtrl is a tool that can potentially be used for mass screening in order to detect traveler that might be suspicious and need to be checked manually by border patrol. 
Sanchez et al. critically analyzed the limited public available information~\cite{sanchez2022politics}.
They found, that the tool uses 38 features, or channels, such as ‘left eye blink’, ‘increase
in face redness’ or head movement directions.
The system was trained on a small, lab condition database where 32 actors play the role of truthful or deceptive travelers, answering 13 questions (such as 'Are there any items from the lists of prohibited items in your case?'). 
 For deception detection, an accuracy of 73.66\% is reported and 75.55\% for detecting truthfulness. 
 However, due to the very limited number of (acted) training samples, Sanchez et al. doubt that the current system is suitable to detect deception in border patrol context without major drawbacks.

Results on the common available deception databases shall be discussed in the next section.

%%%%%%%%%%%%%%%%%%%%%%%%%%%%%%%%%%%%%%%%%
\section{Available Deception Databases}
In the following, we take a closer look at the four publicly available deception detection datasets, three of them have low-stake and one has high-stake context. % namely \ac{bag}, \ac{box}, \ac{mu}, and \ac{rl}.

\begin{figure*}[tb]%
    \centering
    \begin{subfigure}[b]{0.33\textwidth}
       \includegraphics[]{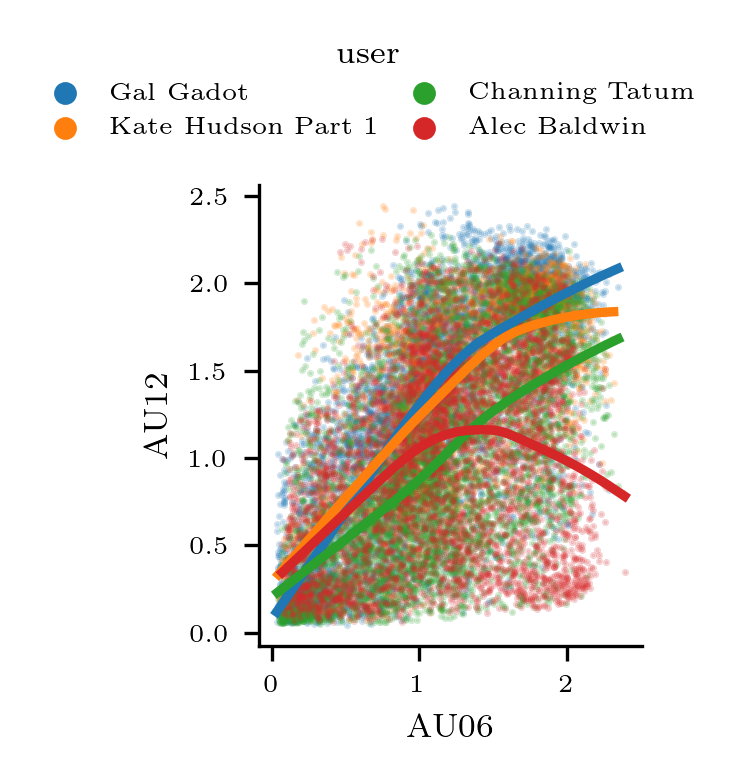}
        \caption{Different subjects from \acl{box}}    
    \end{subfigure}
    %todo: trim y axes: \includegraphics[trim={5mm 0 0 0},clip]    
    \begin{subfigure}[b]{0.33\textwidth}
       \includegraphics[]{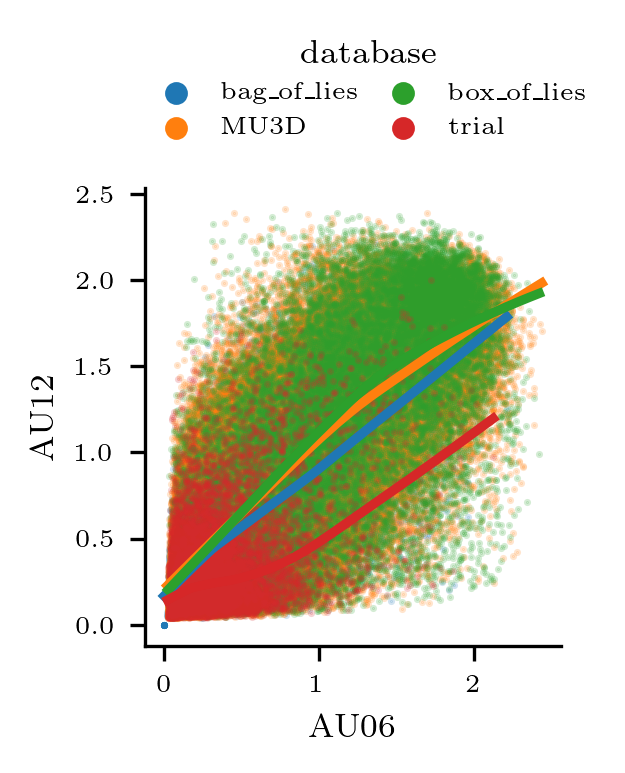}
        \caption{Different datasets}    
    \end{subfigure}
    \hspace{-10mm}%what causes the space here??
    \begin{subfigure}[b]{0.33\textwidth}
       \includegraphics[]{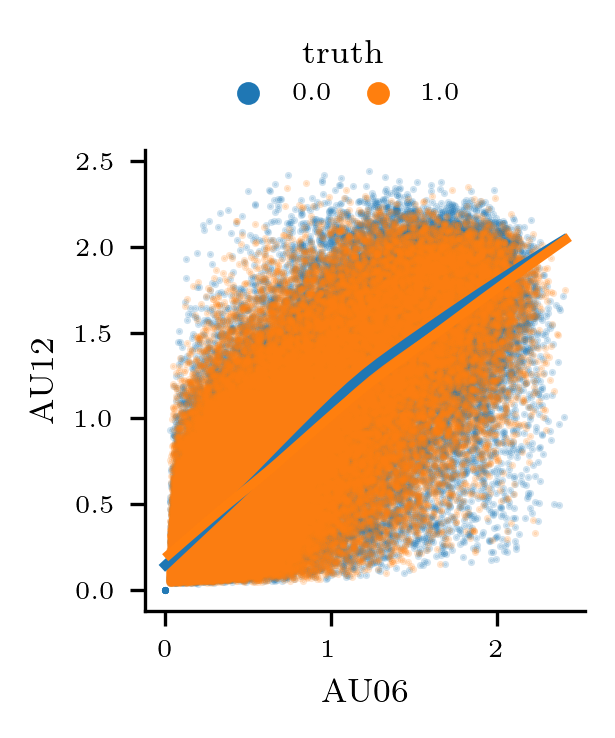}
        \caption{Truth vs deception}    
    \end{subfigure}
     \caption[]{Plot analysis the correlation of the Action Units AU12 (Lip corner puller) and AU06 (Cheek raiser) as indication of a false smile. Deviations can be observed between subjects, databases, however, not between true and lie-samples. }	
    	\label{fig:acted_smile}
\end{figure*}

\subsection{\acl{bag}}
\acf{bag} is a multimodal dataset designed for deception detection using various modalities, including video, audio, and, unlike other datasets, EEG data (for a subset of the database)~\cite{gupta2019bag}. 
The dataset aims to explore the cognitive aspect of deception and combines it with vision, providing a realistic scenario for collecting data.
\ac{bag} contains 35 unique subjects, providing 325 annotated samples with an even distribution of truth and lies. 
The goal of this dataset is to facilitate the development of better deception detection algorithms that are more relevant to real-world scenarios.
The samples are acquired as follows: A photo is displayed on a screen in front of a participant, who is then asked to describe it. The participant can freely choose whether to describe the photo truthfully or with deception. 
A motivation for a deceptive answer, as a higher reward, is not given.

Gupta et al. achieve 66.17 \% Accuracy on a subset of \ac{bag} using EEG, gaze, video, and audio features.
On the full \ac{bag} they got 60.09 \% (without EEG).
Using single modalities, gaze (using an eye tracker system) gives the best accuracy of 61.7\%/57.11\% (on subset/full set) results, while 56.2\%/55.26\% were achieved using video features only.

\subsection{\acl{box}}
\acf{box} is a dataset that is based on the game with the same name that is part of a late night TV show.
In this show, the guest and host take turns describing an object truthful or deceptive (or on the opponent's move guessing whether this description might be the truth) \cite{soldner2019box}.
Besides linguistic and dialog features, also various nonverbal cues are manually ground-truthed over time including features from gaze, eye, mouth, eyebrows, face and head.

Soldner et al. achieve 69\% Accuracy (guests truthful\slash decep\-tive)  training Random-Forest on 60\% of the samples using all the manually extracted features.
Furthermore, the experiments showed that linguistic features were the most relevant (66\% Accuracy), followed by non-verbal cues (61\%) \cite{soldner2019box}. 
Furthermore, accuracy by Human, that is based on guess while watching the video samples, is just 55\%.  
However, during an annotated scene -- which is a time-slot in which the truthful or deceptive description should be given -- cuts are often made, and the camera may shift to the 'opponent' or even both players.
This as well as the fact that \ac{box} is biased (65\% deceptive answers) could be an obstacle using automized feature extraction. 
Apart from that, most guest are actors, which could makes it difficult to transfer any results to real-world scenarios.

\subsubsection{Preliminary Investigation Concerning Fake Smiles}
As a first preliminary investigation, we tried to detect fake smiles using the measured intensities for the \acp{au} AU12 (Lip Corner Pulled) and AU06 (Cheek Raiser).
We expected that AU06 would be low compared to AU12 in case of faked smiles, since they should not involve the eye areas to the same extent.
However, as one can see in Fig. \ref{fig:acted_smile} c), the \acp{au} correlate well for both truthful and deceptive samples.
Nevertheless, there is a clear difference if we compare the four databases: 
 the high-stake \ac{rl} dataset, where the subjects were on trial, shows the lowest values for AU12 compared to the other datasets where AU06 and AU12 show a better correlation (see Fig. \ref{fig:acted_smile} b)).
Furthermore, there is a significant individual influence, as is shown in Fig. \ref{fig:acted_smile} a). 

\begin{figure}[tb!]%
    \centering    
       \includegraphics[]{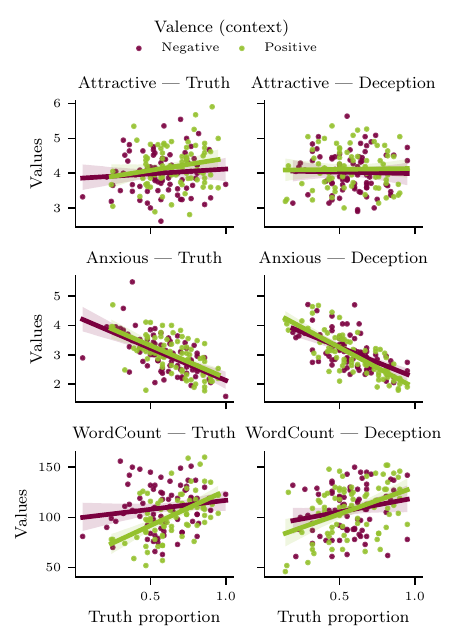}       
     \caption[]{ 
   Three manual features above the truth proportion, separated into positive (green) and negative (red) valence, and deceptive (left) and truthful (right) responses. Each subplot is a scatter plot with linear regression fit where each circle represents a single sample video.
     The truth proportion is the mean value of human raters' belief that the subject answers truthfully in a sample video.      
     }	
    	\label{fig:MU3D_analysis}
\end{figure} 
%\FloatBarrier

\subsection{\acl{mu}}
The \ac{mu} is a free resource that includes 320 videos of individuals telling truths and lies.
The videos feature 80 different targets of different ethnical background.
Each target produced four videos, covering positive and negative truths and lies, which results in a fully crossed dataset to investigate research questions as 'are positive lies more difficult to detect than negative lies?' or how target features as age influence impact deception detection~\cite{lloyd2019miami}.
The videos were transcribed and evaluated by naive raters, and descriptive analyses of the video characteristics and subjective ratings are provided.
The \ac{mu} offers standardized stimuli that can enhance replication among labs, promote the use of signal detection analyses, and facilitate research on the interactive effects of race and gender in deception detection.

\subsubsection{Analysis of \ac{mu}}
A significant feature of \ac{mu}, compared to the other datasets, is the valence ground truth, which is the positive or negative context: telling truth or lies about a person the subject likes or dislikes \cite{lloyd2019miami}.
In the original paper, distributions are provided only for individual features, without exploring their correlation.
Hence, we analyze the truth proportion (guessed by human raters) in relation to some features (which are part of the ground truth) and compare the results for positive and negative valence and for true and deceptive answers, as one can see in Fig. \ref{fig:MU3D_analysis}.

We discovered that as the level of anxiety increased among the subjects, the likelihood of the observers judging them as truthful decreased.
However, we found no significant influence of the actual veracity (truth, deception) or the valence context in this regard. 
This shows firstly that human raters might overestimate the significance of noticed signs of anxiety as a cue for deception, and secondly that the influence of the valence context might not be crucial in case of low-stake scenarios.
However, truth proportion is significantly more sensitive to the word count in case the proband speaks the truth and valence is positive (left bottom subplot of Fig. \ref{fig:MU3D_analysis}). 
Apart from that, the (subjective) attractiveness has only a slight influence on truth proportion in all cases.

     %Features over Truth proportion for positive/negative valence: For example the more anxious the proband is, the less thinks the watcher, that he is honest; this is the case samples where the probands actually lies AND where he tells the truth AND for positive and negative context (no difference beyound the standard error).

\begin{table}[htbp]
\setlength{\tabcolsep}{3.0pt}
\centering
\caption{Comparison of the statistics of the four databases. Here applies $\diameter$: average sample length in seconds and standard deviation, \xmark: No, \cmark: Yes, Unbiased: ratio of truthful and deceptive samples is $\approx0.5$,    ItW: in-the-wild.}
\label{table:databases}
\begin{tabular}{lccccc}
\toprule
Database & Samples & Subj. & $\diameter$  & Unbiased & ItW \\
\midrule
\acs{bag} & 325 & 35 &  9.2$\pm$4.8s & \cmark & \xmark \\
 \acs{box} & 25 & 25 &  154.9$\pm$45.2s & \xmark & \cmark \\
\acs{mu}  & 320 & 82 & 35.7$\pm$3.7s & \cmark & \xmark \\
\acs{rl} & 118 & 56 & 24.9$\pm$14.2s & \cmark & \cmark \\
\acs{dice} & 101 &   101 &   28.31$\pm$2.78 & \xmark & \xmark \\
\bottomrule
\end{tabular}
\end{table}

\subsection{\aclp{rl}}
%The \aclf{rl} dataset...
 % evaluation of the db: "Interestingly, deceivers seem to blink and shake their head less frequently than truth-tellers."
 % "poorest accuracy is obtained in Silent video, followed by Text, Audio, and Full video where the judges have the highest performance" \cite{perez2015deception}

%Automatic deception detection
% -real life trials (manual extracted features, classified with DecisionTrees and RF)
% -leave-one-out cross-validation

  Perez et al. note the importance of identifying deception in court trial data, given the high stakes involved ~\cite{perez2015deception}. In this vein, they introduce \acf{rl} -- a novel dataset consisting of videos from public court trials -- and present a multimodal deception detection system that utilizes both verbal and non-verbal modalities to discriminate between truthful and deceptive statements made by defendants and witnesses. 
  
  The system achieves classification accuracies in the range of 60-75\% using a set of manually extracted features and decision trees or \ac{rf} for classification and leave-one-out cross-validation. This outperforms human capability to detect deception in trial hearings on raw data (62\% accuracy).
  Their findings further indicated that facial features were the most useful for identifying deceit using \ac{ml}.
  Human observers on the other hand were most effective on audio-only or full-video, followed by audio, text and silent video (while agreement of 3 annotators was best for audio-only)
 \cite{perez2015deception}.

\subsubsection{Preliminary Investigation on RL}
\begin{figure*}[t]%
\centering
\includegraphics[]{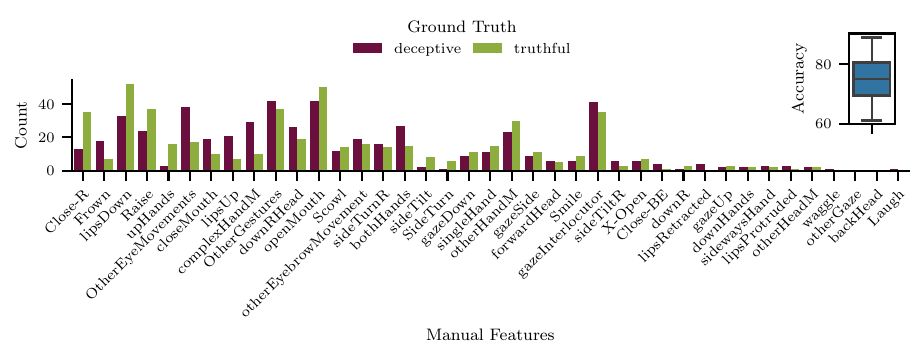}
	\caption[]{
		 Distribution of the given manually labelled features of the \ac{rl} dataset for truth and lie samples. Features are ordered descending from left to right according to the achieved feature relevance based on cross-validation using \ac{svm}.
         On the upper right there is a boxplot including the median of the accuracy which is 77.7\% while average accuracy is 76.2\% (using \ac{rf} we get a median of 75.0\% and average of 74.9\%).
         %Note - Median value: When the median value of classification accuracy is higher than the average accuracy, it means that the distribution of the accuracy values is skewed towards the lower end, with a few outlier values dragging down the average accuracy. In this case, the median value provides a better representation of the typical accuracy value than the average value	   
        }	
	\label{fig:RL_manual}
\end{figure*}

  We repeat the experiment using the given manual features as input for a \ac{svm} using repeated random train-test splits with $k=30$. 
  To further analyze the dataset, we rank the feature according to their relevance and plot their occurrence for samples which are labeled as deceptive respectively truthful, as one can see in Fig. \ref{fig:RL_manual}. Frowning, (eyebrow) raise, or lips down are one of the most relevant features, which are also among the best 6 features reported in ~\cite{perez2015deception}. 
 Using all non-verbal features, we achieve an average accuracy of 76.2\% for SVM respectively 74.9\% for \ac{rf} classification, which is comparable to  Perez et al. which got 73.55\%  using \ac{rf} and 68.59\%  using decision trees ~\cite{perez2015deception}.

%%%%%%%%%%%%%%%%%%%%%%%%%%%%%%%%%%%%%%%%%%%%%%%%%
\section{Methodology}
%Methods/Features that we have used/analysed: SHould we put intermediate-results here or in the exxperimental section??
In this section, we present our approach for deception detection. 
In Section \ref{sec:methodods_cnn} we describe our end-to-end deep learning methods for generating comprehensive features from diverse modalities. Thereafter, we outline our proposed deception classification approaches in section \ref{sec:methodods_classify}.

\subsection{End-to-End learning for High-Level feature generation}\label{sec:methodods_cnn}
In the following, we describe our research towards different modalities, namely head pose, gaze detection and emotion recognition, from which we generate features that are used to detect deceive in a second step.

\subsubsection{Headpose Estimation}
%introduction/problem of current solutions
 Current landmark-free methods often split up continuous rotation variables into bins for classification, leading to a loss of information. 
 %Additionally, using the popular Euler angle or quaternion representations for training neural networks for head pose estimation may not be ideal due to their discontinuity  known as \textit{gimbal lock}: The same visual head pose have infinitive rotation parameterizations.
 Additionally, using the popular Euler angle or quaternion representations for training neural networks for head pose estimation may not be ideal due to their discontinuity, known as \textit{gimbal lock}, for Euler angles, and ambiguity for quaternions.
 %what we did
 To address this, we proposed a landmark-free head pose estimation method in \cite{hempel20226d} %using 
 based on the rotation matrix representation, allowing for full pose regression without ambiguity. This simplifies the network by avoiding performance stabilizing measures used in other methods, such as discretization of rotation variables into a classification problem.
 Instead of predicting the entire nine-parameter rotation matrix, we regress a compressed 6D form, transformed is into the rotation matrix in a subsequent task.
 Furthermore, we use the geodesic loss instead of the often used $l_2$-norm to penalize the network in the training process, capturing the SO(3) manifold geometry. 

%A limitation of the most popular datasets for head pose estimation is, that they cover mainly frontal face samples.
%However, this might be a problem in many real-world  applications, where more extreme head poses might occur.
%This is why we train our model....

\subsubsection{Gaze Estimation}
The majority of CNN-based gaze estimation models predict 3D gaze by regressing the (yaw, pitch) in spherical coordinates and use the mean-squared error for penalizing their networks. We propose a simple network architecture called L2CS-Net~\cite{abdelrahman2022l2cs} based on the ResNet50 backbone and the combination of classification and regression losses. We propose to predict each gaze angle in a separate fully connected layer in order to utilize their independence and to capture the characteristic of each angle by detaching the shared features of the two angles from the last layer of the backbone. Instead of directly predicting continuous gaze angle values directly, we bin the gaze targets to combine classification and regression losses for a coarse-to-fine strategy that effectively promotes our network's gaze performance. 

In addition, we utilize two separate losses, one for each gaze angle. Each loss consists of classification and regression components. We perform gaze classification by utilizing a softmax layer along with cross-entropy loss to obtain coarse gaze direction. On the other hand, we get fine-grained predictions by calculating the expectation of the gaze bin probabilities followed by a gaze regression loss.

%Future: 1. combined gaze&headpose model will faster and improvve gaze
%        2. Mobilenet for faster deception detection

\subsubsection{Micro-Expressions}
%student paper (why we not using ME for lie detection right now)
In a previous experiment, we found that the available \ac{me} datasets were insufficient for proper training
of robust, well generalizing models due to their limited size, uneven \ac{me} class distributions, and an strong overall bias towards certain classes~\cite{talluri2022}.
To address this issue, we combined two databases, using only the three most common ME classes. Although this improved intra-database results (approximately 80\% accuracy), cross-database performance was still not better than random guess.

Considering the limitations of the available datasets, including a lower frame rate observed in all examined deception datasets, we currently do not recommend using \ac{me} for deception detection unless future datasets effectively address these limitations.

\subsubsection{Macro-Expressions}
For macro-expressions we use two models, one to detect \ac{au} intensities and one
to predict expressed emotions directly.

\paragraph{\aclp{au}}
To predict \acp{au} for the experiments of this article,
 we have used a model from Fan et al. that is based on the Resnet50 backbone and was trained on the BP4D database \cite{fan2020fau}.
 Heatmap regression and Semantic Correspondence Convolution were used to achieve reliable predictions of intensities for five different \acp{au}.
 Fan et al. report good results which outperformed 6 other approaches (in average they got an ICC of 0.72 and MAE of 0.58).
For future works, we plan to train our own models on EmotioNet~\cite{fabian2016emotionet}, since it contains 12 \ac{au} classes and covers 100k manually labeled in-the-wild samples.

\paragraph{Multitask Emotion Prediction}
%end-to-end
In addition to using \acp{au}, we can also directly derive emotional states from facial expressions. While it is true that certain \acp{au} can be used to derive emotions, annotating them is a time-consuming process which is why many comprehensive databases do not include \ac{au} labels.
This is why we believe that investigating the use of both \acp{au} and direct prediction of emotions has the potential to yield benefits regarding deception detection.

%AffectNet & Multitask networks
The comprehensive AffectNet database comprises approximately 500,000 samples, each with manual labels for basic emotion categories, as well as values for valence and arousal~\cite{mollahosseini2017affectnet}.
This allows for the simultaneous training of emotion classification models for neutral, happy, sad, surprise, fear, disgust, anger, and contempt (although contempt is considered a secondary emotion).
Additionally, the database allows for the regression of intensities within the range of -1 to 1 for both valence and arousal.

To reduce the computational costs -- which is crucial for our future experiments -- we employ a multitask model to perform both classification and regression with a single network.

%
%\FloatBarrier
\subsection{Deception Detection}\label{sec:methodods_classify}
%end-to-end problematic for deception
The above-described tasks as head pose estimation or emotion prediction described 
above were efficiently solved using end-to-end learning due to the availability of at least one comprehensive dataset.
However, when it comes to deception detection, all available datasets are clearly limited to solving a complex and highly context-sensitive task such as deception detection. 

%transfer learning also not suitable
Transfer learning is not feasible in this case due to the absence of a typical pretraining domain. While macro-expressions exist as a feature of deception, they do not provide a comprehensive and suitable pretraining task. Without a single domain that adequately represents the target task, finding an appropriate base model through transfer learning becomes challenging.
Instead, using multiple modalities, that have been successful in former deception detection research, might be a better and more flexible strategy.
Furthermore, signs of deception are typically ambiguous, very individual, and weakly expressed, especially in low-stake contexts. 
As a matter of fact, end-to-end learning would require even more training data compared to tasks with measurable target values as (headpose) angles, but the datasets cover just a few hundred samples. 
Hence, we use the above-discussed CNN models for head pose and gaze estimation, action unit regression, and classification of basic emotion as well as regression of \ac{va} values as features for deception detection.

\subsubsection{Analysis of the Features}
Each modality consist of several features, as the yaw and pitch angle in case of gaze.
As modalities may not be suitable in the same extent for all datasets - depending on what meaningful patterns there are for these modalities in both the train and also the test set - also the significance of single features may depend on the data.
Hence, we first compare the feature distributions of truthfully and deceptive samples for the four datasets.

\begin{figure}[tb!]%
\centering
\includegraphics[]{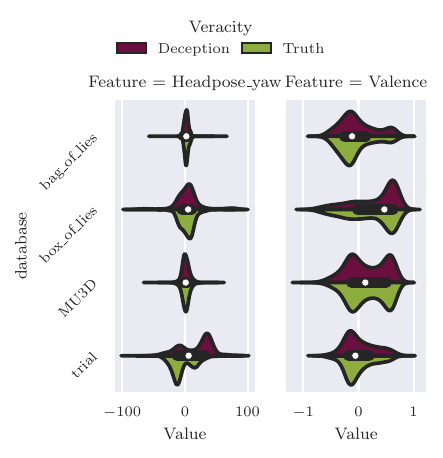}
	\caption[]{Violinplots of key features extracted by end-to-end learning approaches across four publicly available datasets, highlighting correlations with deceptive behavior. }	
	\label{fig:Features_violin}
\end{figure}

As shown in Fig. \ref{fig:Features_violin}, the headpose yaw angle is a very interesting feature in case of the \ac{rl} dataset, since it is clearly higher in case of deception (distribution for gaze yaw angle is quite similar). 
This may be a result of the high-stake context, since a guilty person might tend to avoid eye contact more likely.

\begin{figure*}[tb]%
\centering
\includegraphics[]{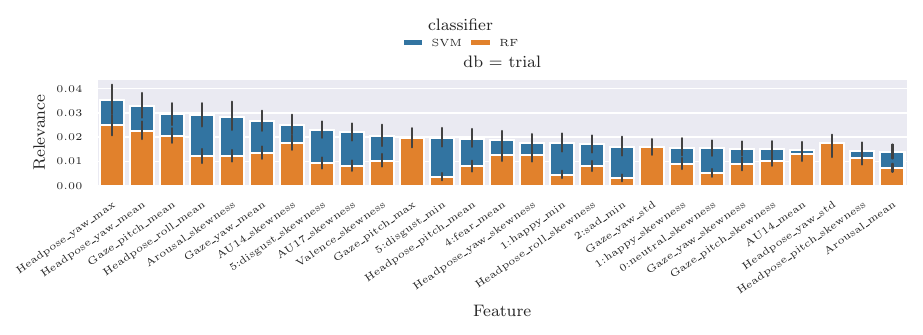}%
	\caption[]{
		 Feature Ranking. We calculated feature relevance through feature permutation using traditional classifiers (SVM and Random Forest) for the \ac{rl} dataset. The resulting list of features was ordered based on their relevance as determined by the SVM classifier, after which we retained the top 25\%. Notably, headpose and gaze features emerged as particularly relevant for the trial database. It is worth noting that the relevance of features can vary depending on the classifier used.		
	}	
	\label{fig:Feature_Ranking_RL_best}
\end{figure*}

In case of the valence feature, the distribution of deceptive and truthful samples is almost identical. 
However, it is obvious that is clearly differing for the different datasets.
Unexpectedly -- despite \ac{bag} being acquired under lab conditions and in a low-stakes context like \ac{mu} -- the valence distribution is predominantly within the negative range,  closely resembling that of \ac{rl} datasets, which typically involve more serious contexts.

The distributions for all used features can be found in Appendix \ref{sec:sample:appendix}.

% derived statistical features for SVM/RF
\subsubsection{Classification using Dicriminative Models}
In the first step of our approach, we employ discriminative models such as \ac{svm} and \ac{rf} to classify the video samples from deception databases as truthful or deceptive. 
To extract meaningful information from each base feature, we derive a range of statistical features such as mean and skewness. We then rank these features based on their relevance. Fig. \ref{fig:Feature_Ranking_RL_best} shows the ranking of the top 25\% of features for both linear SVM and RF for the \ac{rl} dataset. 
As expected, yaw angle of gaze and headpose are among the most relevant features. For a comprehensive ranking of all databases, please refer to Appendix \ref{sec:sample:appendix}.

\paragraph{Feature Selection}
In the next step, we aim to reduce the number of features used in our classification model. 
We have observed that retaining approximately 20-30\% of the most relevant features yields good results and reduces computational costs. 
However, the optimal number of features depends on the dataset and the type of classifier used.

\begin{figure}[tb!]%
\centering
\includegraphics[width =\singlecolumnwidth]{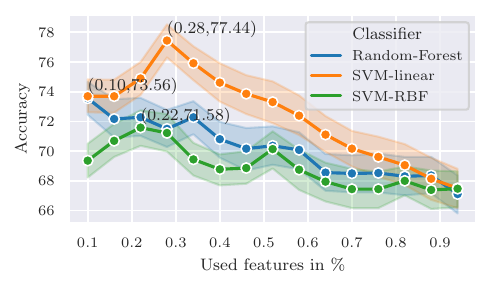}
	\caption[]{Average and standard error (using cross-validation) for different fractions of the most relevant features for the discriminative classifiers.}	
 \label{fig:optimization}
\end{figure}

In Fig. \ref{fig:optimization}, we present the performance of three classifiers on the \ac{rl} dataset. 
Our results indicate that the best accuracy of 77\% is achieved when using approximately 30\% of the best features, while the use of all features results in worse performance for all tested classifiers. 
Nevertheless, the results obtained with all features are still better than those obtained by a trivial classifier or random guess, that both would have 50\% accuracy.

For the \ac{bag} and \ac{box} datasets, we have also found that retaining around 20\% respectively 30\% of the features yields the best performance, while for \ac{mu}, we obtained the best results by keeping approximately 80\% of the features.

We have also investigated the use of principal component analysis (PCA) for feature reduction. However, our results show that PCA does not perform well for our classification task.
This may be due to the non-linear relationships between the features and the outcome, which may be lost after PCA's linear combination, or the sensitivity of PCA to outliers in the data.

%LSTM to use temporal features/pattern
\subsubsection{Classification using Sequence-to-Class Approaches}
Instead of computing statistical features for each sample video, time series analysis can be performed on the data using methods such as LSTM.
Since transformers are designed specifically for problems that have tokens as words as input, they do not work well in case of feature vector input.
This is why we tried different \ac{rnn} architectures including GRU, \acp{lstm} and bidirectional \acp{lstm}, where bidirectional \acp{lstm} performed best.

\subsection{\acl{dice}}
%not everybodies lied, although they can, in our case there is monitoring, also automatically lie detection 
The classical economically motivated \acf{dice} from Fischbacher et al. demonstrated that only 20\% of all individuals who roll a dice without supervision will lie to the fullest extent possible (by claiming to have rolled the highest value when they actually did not), while 39\% will be entirely truthful~\cite{fischbacher2013lies}.
It's worth noting that the participants in this experiment knew that they were not monitored or recorded, so it was impossible to detect lies on an individual level.

In our study, both the participants and the results of the dice rolls are recorded. Participants are informed that the data will be labeled for analysis and research purposes, but that the experimenter will not have access to the records. This is done to ensure the integrity of the data, and only authorized individuals outside the economy research team will have access to the records.

To summarize our experimental setup:

\begin{enumerate}%adapt discription to our experiment
\item We use a single standard six-sided dice, so the possible outcomes are 1, 2, 3, 4, 5, or 6.
\item Reward is 1 Euro times the outcome, except for a result of 6 which results in no reward.
\item We have 100 candidates each roll the dice once. 
\item We analyze the results to determine the frequency of each outcome. 
We also record the claim and the actual dicing by webcam.  
\end{enumerate}

In the experimental section, we will compare the results of our modified rolling dice experiment with those of Fischbacher et al. and, furthermore, test our approach of automatic deception detection on the resulting dataset.

\section{Results and Discussion}
\label{results}
In this section, we present the results and engage in a comprehensive discussion to interpret the findings. In the following subsections, we will first discuss the results of the end-to-end learning approaches for the single modalities. Subsequently, we will evaluate our approach of deception detection built on these results using the four available datasets. 
After examining our new RDE dataset, we will assess the proposed deception detection approach using this dataset and compare the outcomes with the other datasets.

%From Thorstens paper: https://ieeexplore.ieee.org/abstract/document/9897219
\subsection{Results for Headpose estimation}
We utilized Pytorch to implement our proposed network, selecting RepVGG (RepVGG-B1g2) as the backbone model. RepVGG is designed as a multi-branch model for training, and can be converted into a VGG-like architecture for deployment, resulting in the same accuracy but with a shorter inference time. We chose the RepVGG-B1g2 model for its similarity to ResNet50. After testing multiple configurations, we found that a single final layer with 6 output neurons performed the best. The network was trained for 30 epochs using the Adam optimizer with initial learning rates of 1e-5 for the backbone and 1e-4 for the final fully connected layer, with both rates halved every 10 epochs. A batch size of 64 was used.

Our experimental evaluation involved using the synthetic 300W-LP dataset for training and ALFW2000 and BIWI real-world datasets for testing.
Mean Absolute Error (MAE) of Euler angles was used as the evaluation metric, and our rotation matrix predictions were converted into Euler angles. 
As one can see in Table \ref{table:headposeResults}, our method outperformed the current state-of-the-art landmark-free approaches for head pose estimation, achieving almost 20\% lower error rate on the AFLW2000 test dataset and the lowest error rate on all three rotation angles (yaw, pitch, and roll). On the BIWI dataset, our method achieved state-of-the-art results for the MAE. Our approach reported very balanced errors, indicating the network was able to learn consistently and robustly compared to other methods with diverging results on single angle errors.

\begin{table*}[t]
\setlength{\tabcolsep}{4.2pt}
\renewcommand{\arraystretch}{1.1}
\centering
\begin{tabular} { l c@{\hskip .3in} c c c c @{\hskip .5in} c c c c }
\toprule
& & \multicolumn{4}{c}{\textbf{AFLW2000}} {\hskip .5in} & \multicolumn{4}{c}{\textbf{BIWI}}\\
\midrule
  & Full Range$^1$& Yaw & Pitch & Roll & MAE & Yaw & Pitch & Roll & MAE \\
  \midrule
 %HopeNet ($\alpha=2$) ~\cite{Ruiz2018FineGrainedHP}& \xmark  & 6.47 & 6.56 & 5.44 & 6.16  & 5.17& 6.98& 3.39 & 5.18\\ %5.167 6.975 3.388 5.177
 HopeNet ($\alpha=1$) ~\cite{Ruiz2018FineGrainedHP}& \xmark & 6.92 & 6.64 & 5.67 & 6.41 & 4.81& 6.61& 3.27 & 4.90\\ %4.810 6.606 3.269 4.895
 FSA-Net~\cite{Yang_2019_CVPR}& \xmark  & 4.50 & 6.08 & 4.64 & 5.07 & 4.27 & 4.96 & 2.76 & 4.00\\ 
 HPE~\cite{Huang2020ImprovingHP} & \xmark & 4.80 & 6.18 & 4.87 & 5.28 & 3.12 & 5.18 & 4.57 & 4.29 \\
 QuatNet~\cite{8444061}& \xmark  & 3.97 & 5.62  & 3.92 & 4.50 & \textbf{2.94} & 5.49 & 4.01 & 4.15  \\
 WHENet-V~\cite{Zhou2020WHENetRF}& \xmark &4.44& 5.75& 4.31& 4.83 & 3.60 & \textbf{4.10} & 2.73 & 3.48\\
 %WHENet~\cite{Zhou2020WHENetRF} & \cmark \xmark & 5.11 & 6.24 & 4.92 & 5.42 & 3.99 & 4.39 & 3.06 & 3.81 \\
 TriNet~\cite{Cao_2021_WACV}& \cmark  & 4.04 & 5.77 & 4.20 & 4.67 & 4.11 & 4.76 & 3.05 & 3.97 \\
 FDN~\cite{Zhang2020FDNFD} & \xmark & 3.78 & 5.61 & 3.88 & 4.42 & 4.52 & 4.70 & \textbf{2.56} & 3.93\\
 \midrule\\[-2ex] 
 6DRepnet & \cmark &  \textbf{3.63} & \textbf{4.91} & \textbf{3.37} & \textbf{3.97} & 3.24 & 4.48 & 2.68 & \textbf{3.47}\\ 
 \bottomrule
%Epoch _epoch_22.tar: Yaw: 3.6267, Pitch: 4.9066, Roll: 3.3740, MAE: 3.9691 aflw200
%Epoch _epoch_22.tar: Yaw: 3.3522, Pitch: 4.8257, Roll: 2.7265, MAE: 3.6348 biwi
%Epoch _3 Epoch _epoch_3.tar: Yaw: 3.3247, Pitch: 4.4286, Roll: 2.7602, MAE: 3.5045
%Epoch _epoch_10.tar: Yaw: 3.3301, Pitch: 3.9599, Roll: 2.6487, MAE: 3.3129
%Yaw: 3.2408, Pitch: 4.4805, Roll: 2.6800, MAE: 3.4671
\end{tabular}
\caption{Comparisons of our pose estimation with the state-of-the-art methods on the
AFLW2000 and BIWI dataset. All models are trained on the 300W-LP dataset. $^1$ These methods allow full range predictions.}
\label{table:headposeResults}
\end{table*}

\begin{table}[]
\resizebox*{\linewidth}{!}{%
\begin{tabular}{lc}
\toprule
\textbf{Methods}   & \textbf{\begin{tabular}[c]{@{}l@{}}MPIIFaceGaze\end{tabular}} \\ 
\midrule
iTracker (AlexNet) \cite{itracker}              & 5.6\degree  \\
MeNets \cite{MeNets}                            & 4.9\degree  \\
FullFace  (Spatial weights CNN)\cite{fullface} & 4.8\degree  \\
Dilated-Net \cite{dilated}                      & 4.8\degree  \\
RT-Gene(1 model) \cite{rtgene}                  & 4.8\degree  \\
GEDDNet \cite{GEDDNet}                          & 4.5\degree  \\
RT-Gene(4 ensemble) \cite{rtgene}               & 4.3\degree  \\
Bayesian Approach \cite{BayesianApproach}       & 4.3\degree  \\
FAR-Net \cite{farenet}                          & 4.3\degree  \\
CA-Net \cite{ca-net}                            & 4.1\degree  \\
AGE-Net \cite{cvpr2021}                         & 4.09\degree \\ \hline
\textbf{\begin{tabular}[c]{@{}c@{}}L2CS-Net \\ L2CS-Net \end{tabular}} & \textbf{\begin{tabular}[c]{@{}c@{}}3.96\degree\\ 3.92\degree\end{tabular}}                     \\ 
\bottomrule
\end{tabular}%
}
\caption{Comparison of mean angular error between our proposed model and SOTA methods on MPIIGaze dataset}
\label{tab:gaze}
\end{table}

\subsection{Results for Gaze Estimation}
The proposed L2CS-Net for gaze estimation was trained on the MPIIGaze datasets with Adam optimizer and a learning rate of 0.00001 for 50 epochs. We normalize the dataset images as in \cite{zhang2017mpiigaze} to remove the effect of head pose variations. Table~\ref{tab:gaze} shows the comparison of mean angular error between our proposed network and state-of-the-art methods on MPIIFaceGaze dataset. Our proposed network achieves state-of-the-art gaze performance with an angular error of 3.86 on MPIIFaceGaze.

%On Gaze360 dataset, L2CS-Net trained with β = 1 achieved state-of-the-art gaze performance with 10.41◦mean angular error on front 180◦ and 9.02◦ on front-facing, following the same evaluation criteria used in previous related works.

%TODO: merge with new results of next table (Convnext (base) needs to be retrained over weekend) 
\subsection{Results for Emotion Prediction}
 Since the official test set of AffectNet is still not published, we split the available data randomly into 70\% for training and keep the remaining for testing, similar to how Mollahosseini et al. split the data to the original train and test set \cite{mollahosseini2017affectnet}.
 Since AffectNet is relatively comprehensive, the results on our test set should be comparable, but one should note that they may suffer from reduced amount of training samples.
 For training, we use the following augmentation techniques: random crop, random erasing, color fitter, random grayscale and random horizontal flip. Then, the images are normalized to 224$\times$224 pixels.

 %NOTE: MobileOne models have a lower latency even on CPU compared to competing methods. (Review — MobileOne: An Improved One millisecond Mobile Backbone) -> S0 should be as fast as mobinenetv3(small) on CPU
\setlength{\tabcolsep}{8.0pt}
\renewcommand{\arraystretch}{1.1} 
\begin{table*}[tbp!]
	\small
	\caption{Evaluation of our emotion prediction approach on AffectNet}
	\centering
	\begin{threeparttable}
		\begin{tabularx}{1.0\textwidth}
  %{Yp{3.8cm}YYYYYY}
  {X p{4.5cm} c c c c c c}
			\toprule		
			& &  \multicolumn{2}{c}{class} & \multicolumn{2}{c}{valence} &  \multicolumn{2}{c}{arousal}\\
			\cmidrule(r{5pt}){3-4} \cmidrule(r{5pt}){5-6} \cmidrule(r{0pt}){7-8}
			& Approach & Acc & $F_1$ & CCC & RMSE & CCC & RMSE \\
			\midrule
			% ------------ Class Only
			\multirow{5}{*}{\rotatebox{90}{ \cite{mollahosseini2017affectnet}\tnote{1} } } & SVM\tnote{1} & 60 & 0.37 &-&-&-&-\\
			& MSCognitive\tnote{1} & 68 & 0.51 &-&-&-&-\\
			& AlexNet\tnote{1,4} & 64-72 & 0.55-\textbf{0.57} &-&-&-&-\\					
			% ------------ Valence
			%\cmidrule{2-5} 
			& SVR\tnote{1} & - & - & 0.340 & 0.494 & 0.199 & 0.400\\
			& AlexNet\tnote{1} & - & - & 0.541 & 0.394 & 0.450 & 0.402\\
			\midrule
			%
                %& mobileone_s0 & 73.46 & 0.47 &         - &         - &         - &          -  \\
                %& mobileone_s0 &          - &          - & 0.78 & 0.31 & 0.49 & 0.25 \\
                %\midrule
			  \multirow{5}{*}{\rotatebox{90}{\textbf{Proposed multitask\tnote{2}}}}& MobileNet 
  V3 (large) &          70.550 &          0.447 &          0.735 &          0.338 &          0.413 &          0.261  \\
                 & Mobileone (S0) &          73.279 &          0.470 &          0.765 &          0.316 &          0.479 &          0.248 \\
                &    Resnet18 &          71.606 &          0.477 &          0.751 &          0.326 &          0.420 &          0.258  \\
                &   Resnet50d &          74.048 &          0.502 &          0.770 &          0.313 &          0.479 &          0.249  \\
                & EfficientNet V2 (small) &          73.946 &          0.518 & \textbf{0.793} &          0.304 &          0.489 &          0.250  \\
                & ConvNext (pico) &          73.019 &          0.472 &          0.776 &          0.309 &          0.485 &          0.252  \\
                & Swin Transformer\tnote{3} (tiny) &          75.654 &          0.514 &          0.753 &          0.310 &          0.467 &          0.244  \\
                & Swin Transformer\tnote{3} (base) & \textbf{76.813} & 0.535 &          0.782 & \textbf{0.296} & \textbf{0.513} & \textbf{0.238}  \\
			\bottomrule
			%\bottomrule
		\end{tabularx}
		\begin{tablenotes}
			%\item $^1$Trained with class labels only,
                %$^2$trained with valence/arousal values only,  $^3$trained on class labels and valence/arousal simultaneously, 
                \item $^1$\cite{mollahosseini2017affectnet}, trained on full training set and tested on unpublished test set,
                \item $^2$ trained on 70\% of the AffectNet training samples and tested on the remaining (test set still not published), 
                \item $^3$ patch=4, window=7,
                \item $^4$ For CNN based classification, Mollahosseini proposed 4 models with AlexNet backbone 
		\end{tablenotes}
	\end{threeparttable}
	\label{table:eval_affectnet}
\end{table*}

 \subsubsection{Evaluation of backbones for the multitask network}
 The proposed network was trained with the PyTorch framework using Adam optimizer with learning rates from 0.00001 to 0.001 and batch sizes from 8 to 64 depending on the backbone size, and was trained for 50 epochs.
 We use the small validation set of AffectNet to estimate at which epoch the model starts overfitting.
As backbone, we compared various different established but also new models all pretrained on ImageNet (1k).
Results are shown in Table \ref{table:eval_affectnet}, where we also compare with the baseline of Mollahosseini et al. who tested different single task models including \ac{svm} and \ac{svr}.
We achieved best results with a vision transformer (base Swin Transformer), which, however, is also the biggest model that we have tested.

%mobileone
Considering the importance for our future works of fast prediction even on weak hardware, also the 
results for MobileNet and especially the new Mobileone architecture are promising. 
Mobileone is designed to mitigate the architectural and optimization bottlenecks present in other efficient neural networks, making it a highly effective backbone for efficient neural networks on mobile devices \cite{vasu2022improved}.
As one see in  Table \ref{table:eval_affectnet}, the smallest version Mobileone S0 outperforms MobileNet V3 and ResNet18 and almost reaches the performance of ResNet50d.

%EfficientNet
Another interesting architecture is EfficientNet.
EfficientNet is designed to be highly efficient in terms of both computational resources and model size. It achieves state-of-the-art performance on several image classification benchmarks such as ImageNet with significantly fewer parameters and floating-point operations (FLOPs) compared to other state-of-the-art models like ResNet50 \cite{koonce2021efficientnet}. 
When it comes to emotion recognition, this particular method demonstrates exceptional performance in predicting valence, as indicated by its high CCC score. This is particularly relevant because valence provides insight into the degree to which a person is experiencing negative, positive, or neutral emotions.
This is why we use EfficientNet to predict basic emotions and valence arousal values as features for deception detection.

\subsubsection{Classification Results and Relation to \acl{va}}
\begin{figure}[tbp!]%
    \centering   
       \includegraphics[width=\singlecolumnwidth]{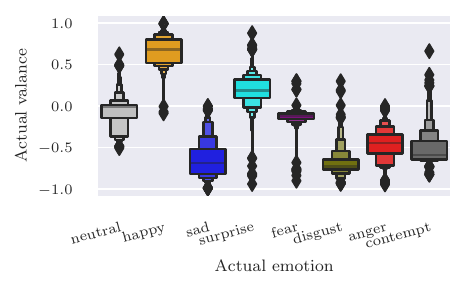}
     \caption[]{Boxenplots of the relation between basic emotions and valence.}	
    	\label{fig:emotion_boxenplots_valence}
\end{figure}

Fig. \ref{fig:emotion_boxenplots_valence} illustrates the relationship between valence and its association with basic emotions, as revealed by the boxenplots that display the quartiles and the range of each emotion's valence values. Happiness is the only emotion that falls distinctly within the positive domain, while surprise and fear are situated close to the midpoint. In contrast, sadness, disgust, anger, and contempt are unequivocally located in the negative territory. 
%contempt probloem
However, valence has a high variation for contempt: Although the median valence value is about -0.5, contempt is often expressed by a wry smile, which would indicate a positive valence and explains the skewed distribution. 
As a matter of fact, it is difficult to proper locate contempt within a \ac{va} circumplex model.

\begin{figure}[tb]%
    \centering   
    %\hspace{-5mm}
    % \begin{subfigure}[b]{0.48\textwidth}
       \includegraphics[width=\singlecolumnwidth]{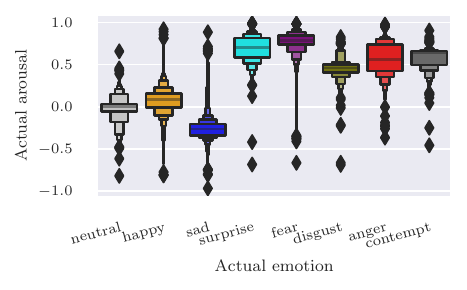}
    % \caption{Arousal}    
    %\end{subfigure}
      %
     \caption[]{Boxenplots of the relation between basic emotions and arousal.}	
    	\label{fig:emotion_boxenplots_arousal}
\end{figure}

In case of arousal, happy is close the midpoint and sad slightly negative, while all other emotions have a median value of at least 0.5, as shown in Figure Fig. \ref{fig:emotion_boxenplots_arousal}.
Detecting very low arousal states, such as tranquility, boredom, or tiredness, can be challenging using AffectNet, as this dataset includes only a limited number of such samples. Nevertheless, we believe that arousal values $\in [0,1]$ are more informative in the context of deception detection anyway. This range of values may be more expressive of subtle changes in arousal, and could help to identify deceptive respectively truthful responses that involve a suppression or concealment of emotional reactions.

It's worth noting that the distribution depicted in Fig. \ref{fig:emotion_boxenplots_valence} and \ref{fig:emotion_boxenplots_arousal}  reflects the \ac{va} circumplex model used to annotate AffectNet. It's important to acknowledge that, unlike the concept of basic emotions, the \ac{va} circumplex model can vary across different datasets. For instance, the circumplex model used in the AffWild database differs from the one used in AffectNet.

\begin{figure}[tbph!]%
\centering
\includegraphics[width=\singlecolumnwidth]{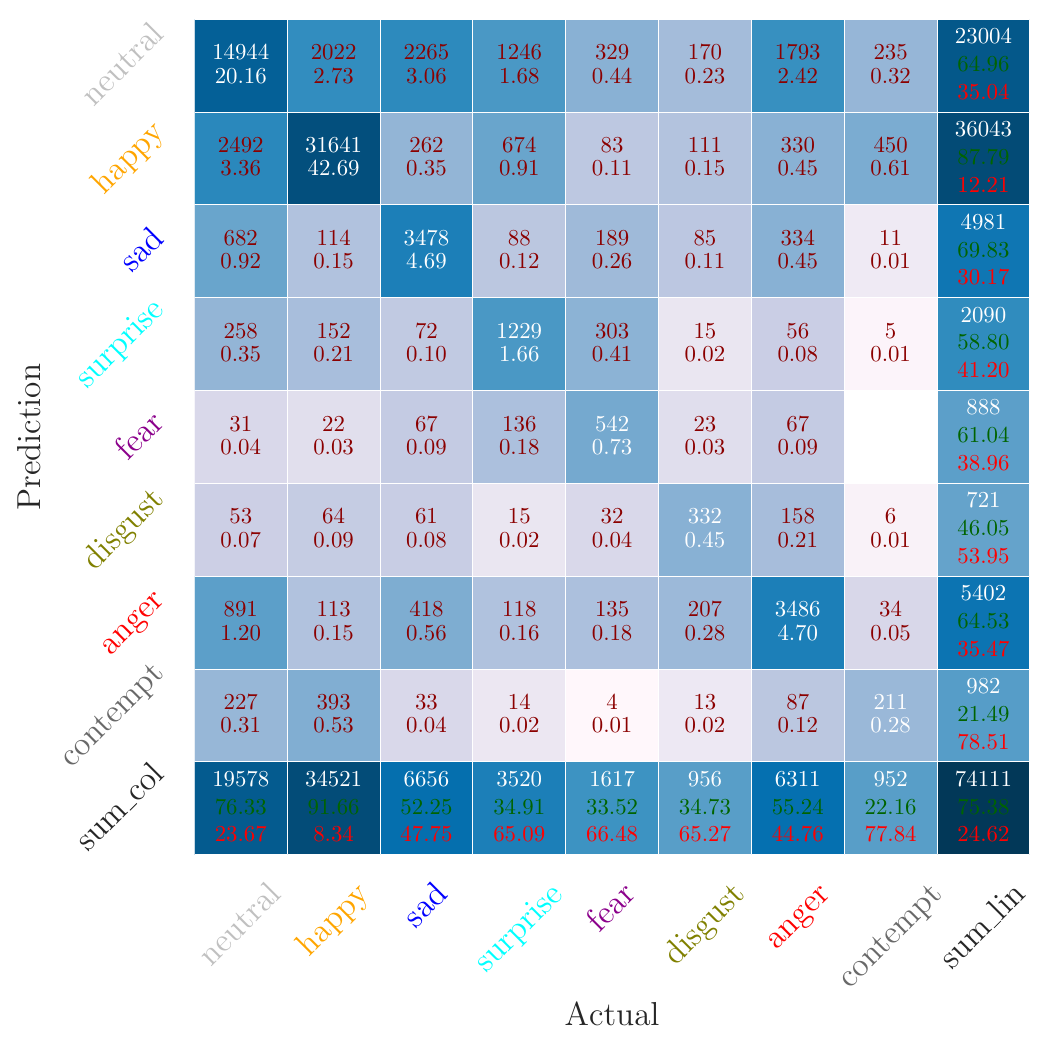}%
	\caption[]{
		 Confusion Matrix for the basic emotions (using a Vision-Transformer).		
	}	
	\label{fig:emotion_confusion}
\end{figure}

\paragraph{Emotion Class Confusion}
In Figure \ref{fig:emotion_confusion}, the confusion matrix shows detailed classification results of the base Swin Transformer -- the tested model with the best overall performance -- on the AffectNet test set.
Although contempt may be more meaningful than the other classes in the context of deception detection, its high confusion rate with other emotions renders it a less reliable feature for this purpose. 
Possible explanations for the high confusion between contempt and neutral or happy emotions is that the facial expressions for contempt can be quite similar to happy, and there are only a limited number of samples available for the contempt class.

Apart from the problems regarding contempt, the confusion matrix reveals another issue.
AffectNet includes numerous instances where annotators had disagreements regarding the appropriate class label, potentially resulting in misclassification. This is particularly evident in cases where the facial expressions are subtle, making it difficult to differentiate between neutral and emotional states. 
On the other hand, such subtle facial expressions are important for real life scenarios, where emotion can be mixed or of weak intensity.
This highlights the significance of incorporating \ac{va} values as features for deception detection, which capture the intensity of the displayed emotions, in order to minimize confusion with neutral expressions.

\subsection{Evaluation of the Four Available Deception Detection Datasets}
In this section, we will conduct experimental evaluations on the given four databases using discriminative classifiers and \ac{lstm} models. These evaluations will be performed using multimodal features extracted from the deep learning approaches described earlier.
Furthermore, we will proceed to conduct additional experiments using the data from our \acf{dice}.

\subsubsection{\ac{lstm}}
\begin{figure*}[tb!]%
\centering
\includegraphics[width=\textwidth]{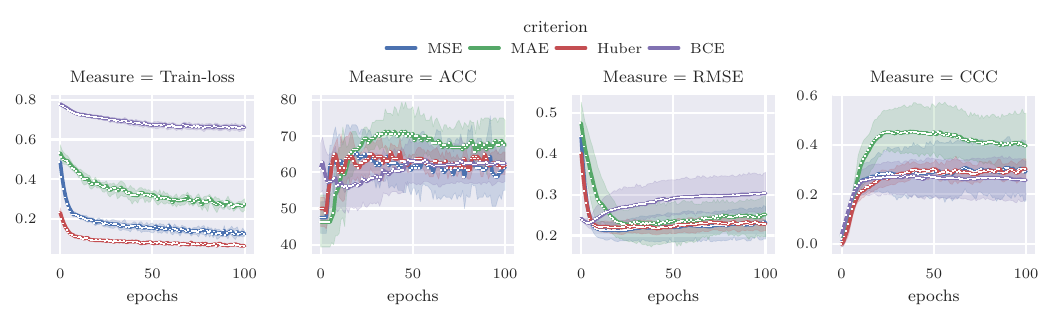}
	\caption[]{Results of the LSTM training on the trial database.}	
	\label{fig:lstm_training}
\end{figure*}

%Reason and problems for LSTM
It is important to note that approaches like \ac{lstm} may struggle to learn meaningful patterns if the available datasets lack comprehensiveness. However, given the possibility of a temporal pattern existing in at least one of the datasets, which could indicate deception, we train \acp{lstm} using the most prevalent criteria. This allows us to evaluate whether \ac{lstm} performs better or worse compared to discriminative models in terms of detection accuracy.
For the \ac{rl} dataset, Fig. \ref{fig:lstm_training} shows the training losses of the used criteria and the validation results for different measures.
%MAE is best
It is evident that Mean Absolute Error (MAE) outperforms other loss functions such as Binary Cross Entropy (BCE) in terms of the evaluation metrics Accuracy and CCC (Concordance Correlation Coefficient). Despite initial expectations favoring BCE as the superior loss function, the results indicate that MAE yields better performance according to these evaluation metrics.

MAE loss is computed as the average absolute difference between predicted and actual values. In the context of binary classification, the predicted values represent probabilities or scores indicating the likelihood of belonging to a specific class.
One of the reasons why MAE performs well could be its robustness to outliers, unlike Mean Squared Error (MSE) or BCE loss functions. Considering the challenging nature of the problem and the limitations of the dataset, which may lead to an increased number of outliers, the resilience of MAE loss to these extreme values becomes advantageous. Furthermore, MAE loss equally considers both overestimations and underestimations, a crucial characteristic for accurate binary classification.

Overall, the superior performance of MAE over other measures suggests its suitability for the given binary classification task, highlighting its ability to handle outliers effectively and treat overestimations and underestimations equally.
Nevertheless, we will see that discriminative models still outperform \ac{lstm} on the deception datasets.

%maybe no seperate section for lstm
\subsubsection{Cross- and Intra-database Results}
%Intra and Crossdatabase results with 4 classifiers
As mentioned earlier, the identification of deception is intricately tied to the specific problem being addressed. However, as far as our knowledge extends, no cross-database experiments have been conducted to assess this aspect qualitatively across the four publicly available deception databases.

\begin{figure*}[tb]%
\centering
\includegraphics[]{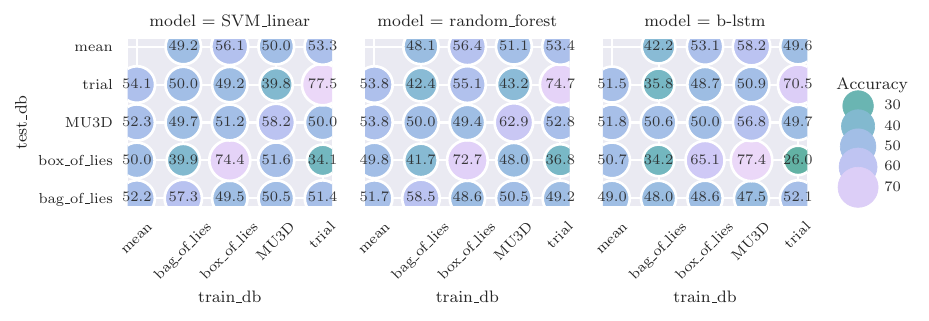}
	\caption[]{
		 Results for Intra- and Cross-database experiments for three different classifiers. 		
	}	
	\label{fig:Crossdatabase_table}
\end{figure*}

Fig. \ref{fig:Crossdatabase_table} shows the cross and intra-database results for \ac{svm}, \ac{rf} and \ac{lstm}.
Cross-database results were obtained by training the model on all samples from one database and evaluating it on all samples from another.
Conversely, intra-database results were obtained using cross-validation, specifically the Repeated Random Train-Test Splits method with a total of 50 iterations for \ac{svm} and \ac{rf} and 10 iterations for \ac{lstm}, utilizing 70\% of the samples for training purposes.
 
 %CROSS
 \paragraph{Cross-database Results} 
 As evident from the results in Fig. \ref{fig:Crossdatabase_table}, most cross-database experiments yield poor outcomes, sometimes even worse than random guess.

 % NOTE: LSTM is best for box of lies. Why (MU3d has more samples and LSTM would need more samples than SVM)
%  -> Maybe there are more relevant temporal features or LSTM profits more from unbalanced data?

In case of the \ac{box} dataset, the underperformance can be attributed to the imbalanced distribution of truthful and deceptive samples when utilizing \ac{box} as a training or test dataset. 
Moreover, the \ac{box} database contains numerous surprisingly short true samples (1 or 2 seconds), necessitating a sequence length of $\leq$ 40 for \ac{lstm} models, whereas the other datasets perform best with a length of 200.
In fact, when employing \ac{lstm} as a classifier in cross-database experiments, the excessively small samples from \ac{box} must be excluded. This exclusion further exacerbates the dataset's imbalance.
This explains the relatively good results using \ac{mu} for training, \ac{box} for testing and \ac{lstm} as classifier, where the model actually does not perform better than a trivial classifier.
% the training dataset gives no information about the majority class of the biased test dataset
%Furthermore, in the case of using \ac{rl} for training, accuracy is clearly beyond the 50\% but on the other hand close to 50\% if both the training and testing dataset is unbiased.

Additionally, it is worth noting that the contribution of modalities varies across databases, which is another contributing factor to the unsatisfactory performance of cross-database results.
This highlights the importance of training deception detection models on datasets that are specifically tailored to the target scenario, emphasizing that the choice of the dataset is even more critical than in general machine learning problems.

\paragraph{Intra-database Results}
 %intradataset and different classifiers
 The intra-database results, provide insight into the performance of the tested classifiers across the four datasets. 

 In general, the \ac{box} dataset achieves the highest accuracy.
 However, it should be noted that this dataset is biased, which leads to classification results benefiting from a priori knowledge. 
 For instance, a trivial classifier achieves an accuracy of 64\% instead of the expected 50\% for the \ac{rl} and \ac{mu} datasets, and 52\% for the \ac{bag} dataset.
Specifically, the accuracy rates on \ac{box} are 74\% for \ac{svm}, 72\% for \ac{rf}, and 65\% for \ac{lstm}. However, it is crucial to consider the F1-scores, as they are less sensitive to biased class distributions. The F1-scores on \ac{box} are 0.57, 0.60, and 0.0/NaN for \ac{svm}, \ac{rf} and \ac{lstm} classifiers respectively. 
In the case of a F1-score of 0, it indicates that either the recall or sensitivity is 0, resulting in an undefined value (NaN). This demonstrates that the \ac{lstm} model fails to identify any significant features that surpass the reliability of a priori information, while \ac{svm} and \ac{rf} exhibit the ability to do so, even with limited training data available.

\subsection{Evaluation of our \acl{dice}}
%economical aspects (impact of monitoring on lie behavior)
In addition to the cross-dataset analysis presented in the previous section, we further examined the performance of our proposed approach by investigating single modalities, alternative measures, and incorporating data from our \ac{dice}.

\begin{figure}[t]%
\centering
\includegraphics[width=\singlecolumnwidth]{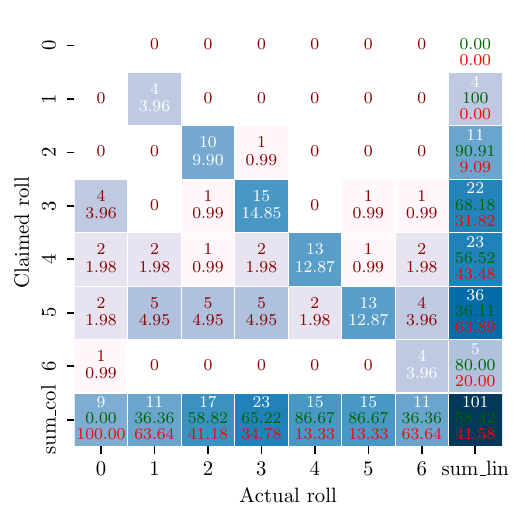}%
	\caption[]{
		The confusion matrix illustrates the subjects' reported dice rolls after actually rolling a specific value between 1 and 6, or not rolling at all.		
	}	
	\label{fig:rde_confusion}
\end{figure}

First, let's examine the \ac{dice} from an economic perspective. Fig. \ref{fig:rde_confusion} depicts the confusion between the actual and claimed values in our \ac{dice}, where participants were aware of being recorded and the records were subsequently analyzed by an external entity, separate from the experiment-conducting institution.

Notably, the percentage of subjects who were completely truthful increased to 58\%, as compared to the 39\% reported in the earlier study conducted by Fischbacher et al. without the presence of records \cite{fischbacher2013lies}. This suggests that the documented record of deceptive behavior had a noticeable yet moderate impact on the participants' behavior.

Furthermore, the findings indicate a strong correlation between the actual value of the dice roll and the participants' behavior. Specifically, most individuals who rolled a 1 or 6 (representing the two worst possible outcomes) claimed they rolled a 4 or 5 (the two best possible outcomes).
Conversely, most participants who rolled a 2, 3, 4, or 5 were truthful about their roll. 
In the end, out of the 101 subjects, 27 falsely claimed to have achieved a better roll than they actually had, while 6 subjects claimed to have rolled worse than their actual result, and 59 called the actual value.
Additionally, 9 participants did not roll at all (indicated with 0).
These discoveries suggest that the accuracy of self-reported data in experiments can be influenced by various factors, including the presence of a record that exposes deception and the actual value of the outcome. 

Please note that the recording also enables a more accurate calculation of the overall deception rate. 
If -- like in \cite{fischbacher2013lies} -- no knowledge of the actual dice roll is available, the claimed roll can only be compared to a statistical distribution (which is an equal probability distribution in the case of a fair dice). However, using only a hundred or a few hundred samples, the actual distribution can clearly differ from the expected one, as it is the case in Fig. \ref{fig:rde_confusion}.
Using 100 rolls, the average deviation from the ideal distribution is $18.24 \pm 5.2\%$, where $\pm$ indicates the standard deviation for simulating the experiment 50 times.
Fischbacher et al. conducted their main experiment with 389 rolls~\cite{fischbacher2013lies}, resulting in a deviation of $8.6 \pm 3.1\%$. 
In our case, this deviation has no influence on the calculated deception rate, since we use the recorded actual rolls as ground truth.

\subsubsection{RoC Curves and F1-scores}
\begin{figure*}[t]%
\centering
\includegraphics[]{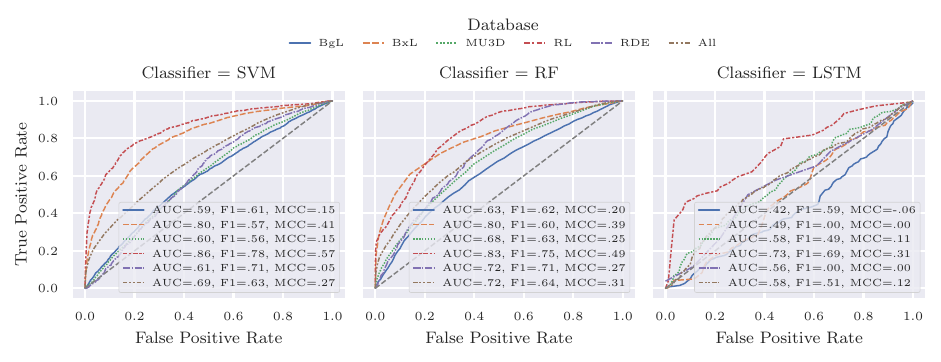}
	\caption[]{
Evaluation of the performance of classifiers using ROC curves, along with additional measures such as AUC and F1-Score, on various datasets. The ROC curves helped us assess the trade-off between true positive rate and false positive rate at different classification thresholds. For this experiment, deception is used as positive class.	  		
	}	
	\label{fig:RoC}
\end{figure*}

To evaluate our deception detection approach on the \ac{dice} and to compare the results with the four available datasets, we compute \ac{roc} curves, F1-scores and MCC measure for all three classifiers as shown in Fig. \ref{fig:RoC}.

RoC curves are advantageous for evaluating model performance in binary classification problems with multiple datasets. They provide a visual representation of the trade-off between true positive and false positive rates, making them useful for comparing models across datasets with varying biases. The derived Area Under the Curve (AUC) measure allows comparing RoC curves of different models with a single scalar. On the other hand, F1-score combines precision and recall into a single metric, providing a balanced evaluation of the model's performance. MCC on the other hand takes into account true negatives and is more suitable for imbalanced datasets, as \ac{box} and \ac{dice}. Utilizing RoC curves and F1-scores and MCC helps assess model effectiveness in the presence of biased and unbiased datasets.

It is evident that \acp{lstm} perform worst over all in terms of AUC, F1-Score and MCC.   
In the sense of AUC, this is particularly true for \ac{bag} and \ac{box}, while F1-Scores and MCC are 0 for the biased \ac{box} and \ac{dice}.
Although on \ac{bag} an F1 score of 0.59 is achieved, the AUC is less than 0.5 and the MCC is slightly negative, indicating that the classifier did not successfully learn meaningful features or that these features are not present in the test set. 
When comparing the AUC and MCC scores of all four low-stake datasets, it is observed that the \ac{lstm} model performs the best on the \ac{mu} dataset. This result was expected since the \ac{mu} dataset has the highest number of samples.

Results on the high-stake \ac{rl} dataset are better, but still not good compared to 
the discriminative classifiers. \ac{svm} achieves the best results on \ac{rl} for all measures, while \ac{rf} exhibits a significantly higher AUC and MCC on \ac{dice}, indicating that \ac{rf} exhibits a better ability to capture the overall performance of the classification task on this more challenging dataset.
%, \ac{rf} achieves an AUC of 0.72, F1 of 0.64, and MCC of 0.31 whereas \ac{svm} attains an AUC of 0.69, F1 %of 0.63 and MCC of 0.27. Therefore,
Moreover, when considering the results using data from all five datasets collectively,
overall both discriminative classifiers yield comparable results, but \ac{rf} may be slightly more accurate, 
as shown in Fig. \ref{fig:RoC}.

In conclusion, while LSTMs are often regarded as more sophisticated models compared to discriminative classifiers, their limited suitability for our problem arises due to the insufficient amount of available data.
Considering the challenging task of deception detection without control questions and contact-based sensors, results are promising, especially for the high stake \ac{rl} dataset. 
In case of our \ac{dice}, detecting deception is much harder, as expected due to the low-stake context. However, compared to \ac{bag} and \ac{mu}, results are good in sense of AUC, F1-score and MCC using \ac{rf} as classifier.

\subsubsection{Evaluation of Single Modalities}
In this section, we examine the effectiveness of each modality in distinguishing between deception and truth. Although utilizing all available modalities, including specialized sensors like thermal data or contact-based modalities such as skin conductance or EEG, can be costly, impractical, or prone to errors in many applications, employing multiple modalities from a single device such as webcams still offers advantages for demanding classification tasks like deception detection.

Table \ref{table:modalities} shows results for single and combined modalities for all datasets using \ac{svm} and \ac{rf} (as shown above, \ac{lstm} is not working well on the limited number of samples, so we skip it for this experiment).
As one can see, the performance of the modalities depends on the used dataset. However, the emotion modality is best for three datasets using \ac{svm} and for two datasets using \ac{rf}. Furtheremore, the best modality is the same for \ac{svm} and \ac{rf} for three of the five datasets.
Apart from that, using all modalities, \ac{svm} outperforms \ac{rf} on three database. However, on the relative comprehensive \ac{mu} and \ac{dice}, \ac{rf} performs better. In both cases, the emotion modality is clearly more effective than the other modalities, especially for the \ac{dice}, where results are even better than using all modalities.
Analysing the contribution of the base features, we found that especially fear, disgust and arousal are relevant.

Table \ref{table:modalities} presents the results for single and combined modalities across all datasets using \ac{svm} and \ac{rf}. As the last experiments shows that -- due to the limited number of samples -- \ac{lstm} did not yield satisfactory performance, it was excluded from this experiment.

It is evident from the table that the performance of the modalities varies depending on the dataset. Notably, the emotion modality demonstrates the highest performance in three datasets when using \ac{svm}, and in two datasets when using \ac{rf}. Additionally, in three out of the five datasets, the same modality performs best for both \ac{svm} and \ac{rf}.

Furthermore, when considering all modalities, \ac{svm} outperforms \ac{rf} in three databases. However, \ac{rf} exhibits superior performance on the relatively comprehensive \ac{mu} dataset and our \ac{dice}. In both cases, the emotion modality proves to be significantly more effective than the other modalities, particularly in terms of \ac{dice}, where its results surpass those achieved by using all modalities, although results are barely above those of the trivial classifier in case of \ac{svm}.

Overall, it can be concluded that the emotion modality exhibits the highest potential. However, it is reasonable to assume that the recognition rates could be further enhanced by incorporating additional modalities, such as those based on audio sensors. Non-contact vital parameter estimation, such as monitoring heart rate or changes in blood circulation over time, may also prove valuable, although it may encounter limitations in certain scenarios and is very sensitive. These limitations include very short samples, variations in lighting conditions, and especially video compression, which can impact the functionality of such modalities.

\begin{figure*}[t]%
\centering
\includegraphics[width=\textwidth]{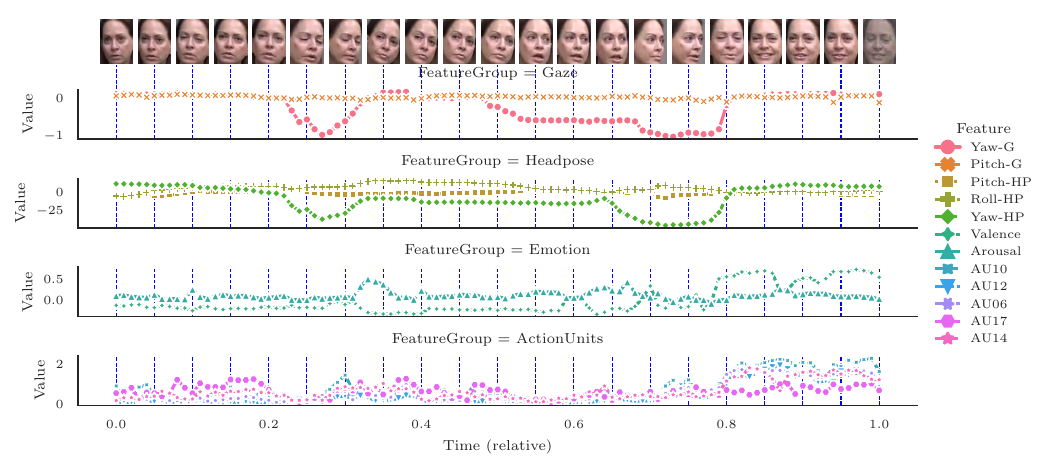}
	\caption[]{
		 Qualitative result shows the most relevant features over time for sample 47 of the lie set of the \ac{rl} dataset (G=gaze in RAD, HP=head pose in $\degree$). 		
	}	
	\label{fig:Qualitative}
\end{figure*} 

\subsubsection{Qualitative Results}
Finally, Fig. \ref{fig:Qualitative} presents a qualitative result from a single sample of the RL dataset. The figure displays the basic features of all modalities over time, along with snapshots of the video data taken at regular intervals (basic emotions are omitted for clarity).

In this particular sample, where the subject exhibits deceptive behavior, there is an observed increase in valence during the last quarter of the video. This observation can be seen as an example of "liar's delight," wherein individuals derive a sense of satisfaction or enjoyment from engaging in deceptive behavior.

Additionally, there is a noticeable change in yaw of gaze and head pose, suggesting that the subject may be intentionally avoiding eye contact with the judge or lawyer.
While the assumption concerning the subject's motivation is challenging to verify definitively, our prior findings highlight the increased relevance of gaze and head pose within the RL dataset compared to other datasets. This observation lends support to the hypothesis that the subject's avoidance of eye contact could be associated with deceptive behavior.

Please note that this qualitative result serves as an illustrative example, showcasing intriguing behavioral patterns and features over time. It is important to acknowledge that not all samples exhibit such pronounced behavior, particularly in the case of low-stake datasets. Extracting indicative features for deception in average examples may present a greater challenge.

\begin{table}
\setlength{\tabcolsep}{3pt}
\renewcommand{\arraystretch}{1.1} 
\centering
\caption{Accuracy of each feature group for linear SVM and Random Forest. Highlighted are the best feature for each database seperately for each classifier and the result for the best classifier for each database in case of all feature groups.}
\begin{tabular}{lcrrrrr}
\toprule
 & feature group &  \acs{bag} &  \acs{box} &  \acs{mu} &  \acs{rl} & \acs{dice}\\
\midrule
        \multirow{5}{*}{\rotatebox{90}{SVM}} &   gaze &      \textbf{54.63} &        69.18 & 52.48 &  68.00 & 58.06 \\
        &   AU &        52.02 &        65.88 & 50.51 &  64.56 & 58.06 \\
        & pose &        51.96 &        64.19 & 53.04 &  \textbf{71.06} & 58.03 \\
      & emotion &        53.32 &        \textbf{69.48} & \textbf{53.15} &  68.44 &  \textbf{59.74} \\         
      \cmidrule{2-7}
      &    all &         \textbf{57.16} &         \textbf{74.33} & 58.25 &   \textbf{77.44} & 62.90\\
          %%RF
           \midrule
           \midrule
      \multirow{5}{*}{\rotatebox{90}{Random Forest}} &       gaze &        55.66 &        70.33 & 57.00 &  70.39  & 54.77\\
          & AU &        52.21 &         \textbf{71.88} & 56.81 &  66.67 & 56.84 \\
        & pose &         \textbf{56.15} &        67.25 & 55.42 &   \textbf{71.61} & 59.68\\
      & emotion &        53.85 &        70.12 &  \textbf{61.81} &  67.61 & \textbf{69.48} \\
        \cmidrule{2-7}
        &  all &        55.18 &        73.22 &  \textbf{63.17} &  73.56 & \textbf{67.96} \\      
       \midrule
       \multicolumn{2}{c}{trivial classifier} &  51.60 & 64.20 & 50.00 & 50.00 & 58.06 \\
\bottomrule
\label{table:modalities}
\end{tabular}
\end{table}

\section{Conclusion}
In this study, we conducted an analysis and comparison of four well-known deception datasets that are publicly available. Considering the limited number of training samples within these datasets, we first developed and evaluated approaches utilizing end-to-end learning to compute intermediate results for single modalities, such as head pose, which are then utilized as features for discriminative approaches. This procedure proves to be more effective when dealing with limited samples.

We evaluated our approach on all datasets and observed that different modalities show varying effectiveness across the different datasets. Additionally, the unsatisfactory results obtained in cross-dataset experiments highlight the high sensitivity of deception detection to specific scenarios.
We conducted an analysis of the features used across different databases and discovered that the distribution of these features varied among the datasets, supporting our initial assumption.
Moreover, we discovered that detecting deception is considerably more challenging in low-stake contexts compared to high-stake contexts. 

Furthermore, we performed an \acf{dice}, generating a dataset with an low-stake economical context.
Previously, subjects were not recorded in the existing \acp{dice}. 
We found, that recording subjects mildly increases the percentage of honest subjects.
More important, recording enables the evaluation of the proposed deception detection approach. We observed that, similar to other low-stake datasets, deception detection in this scenario is challenging due to limited signs of e.g. guilt. However, by employing the appropriate combination of modalities and classifiers, we achieved an average accuracy of 67\% and an F1-score of 0.71. 
Incorporating additional modalities has the potential to enhance accuracy. It is important to note, however, that achieving very high accuracies, such as the claimed 90\% accuracy in polygraph tests, is unlikely in fully automated low-stake scenarios that do not involve contact-based modalities or interrogation techniques as control questions.

We are currently in the process of acquiring a more comprehensive database consisting of approximately 1000 samples. This database will include short sequences depicting ``salespeople''-subjects engaging in simulated online meetings to sell products of either good or of poor quality to customers.
The customers, in turn, must decide whether to trust the salespeople and make a purchase or not.
In addition to the \ac{dice} proposed in this paper, this new dataset will be developed where subjects will actively deceive other participants, providing a unique opportunity to study deception in a more dynamic and realistic setting.
Furthermore, the subjects aim to target a higher payoff of 30 Euros. With the increased number of samples and the introduction of active deception among subjects, we anticipate that this new dataset will enable us to train a more robust deception detection model specifically tailored to this scenario. 

Looking ahead, our future research will focus on optimizing our proposed approach within this specific context and seamlessly integrating deception detection directly into the experiment. By continuing to explore and refine our methods, we aim to contribute to the advancement of accurate and effective deception detection techniques.
This also encompasses the investigation of additional modalities, such as audio, vital parameters, and, where applicable, linguistic analysis of the spoken language.

\backmatter

\bmhead{Supplementary Information}
Not applicable

\bmhead{Funding}
Open Access funding enabled and organized by Projekt
DEAL. 
This research was funded by the Federal Ministry of Education and Research of Germany (BMBF) project AutoKoWaT, no. 13N16336 and by the German Research Foundation (DFG) project AL 638/13-1, AL 638/14-1 and AL 638/15-1.
%Additionally, we acknowledge support by the Open Access Publication Fund of Magdeburg University.

\bmhead{Data Availability}
The publicly available datasets used for this research are described in ~\cite{gupta2019bag},~\cite{soldner2019box},~\cite{lloyd2019miami} and ~\cite{perez2015deception}.
The \ac{dice} dataset can not be published due to data protection reasons.

\section*{Declarations}

\bmhead{Conflict of Interest}
The authors declare that they have no conflict of interest.

\bmhead{CrediT Authorship Contribution}
 Conceptualization:  Laslo Dinges, Marc Fiedler; Methodology: Laslo Dinges; Software: Laslo Dinges, Thorsten Hempel, Ahmed Abdelrahmann; Dataset acquisition and labeling: Joachim Weimann, Dmitri Bershadskyy, Laslo Dinges, Ayoub Al-Hamadi; Project heading: Joachim Weimann, Ayoub Al-Hamadi; Funding acquisition: Ayoub Al-Hamadi; Writing - original draft preparation: Laslo Dinges; Writing - review and editing: Laslo Dinges, Marc Fiedler, Dmitri Bershadskyy; Supervision: Ayoub Al-Hamadi 

\begin{appendices}

\section{Feature distribution and relevance}\label{sec:sample:appendix}
In this section, we present extended plots that provide an overview of multiple datasets and all features. While these plots may not be essential for understanding the main paper, they offer additional insights and details for interested readers.

\begin{figure*}[p!]%
\centering
\includegraphics[width=\textwidth]{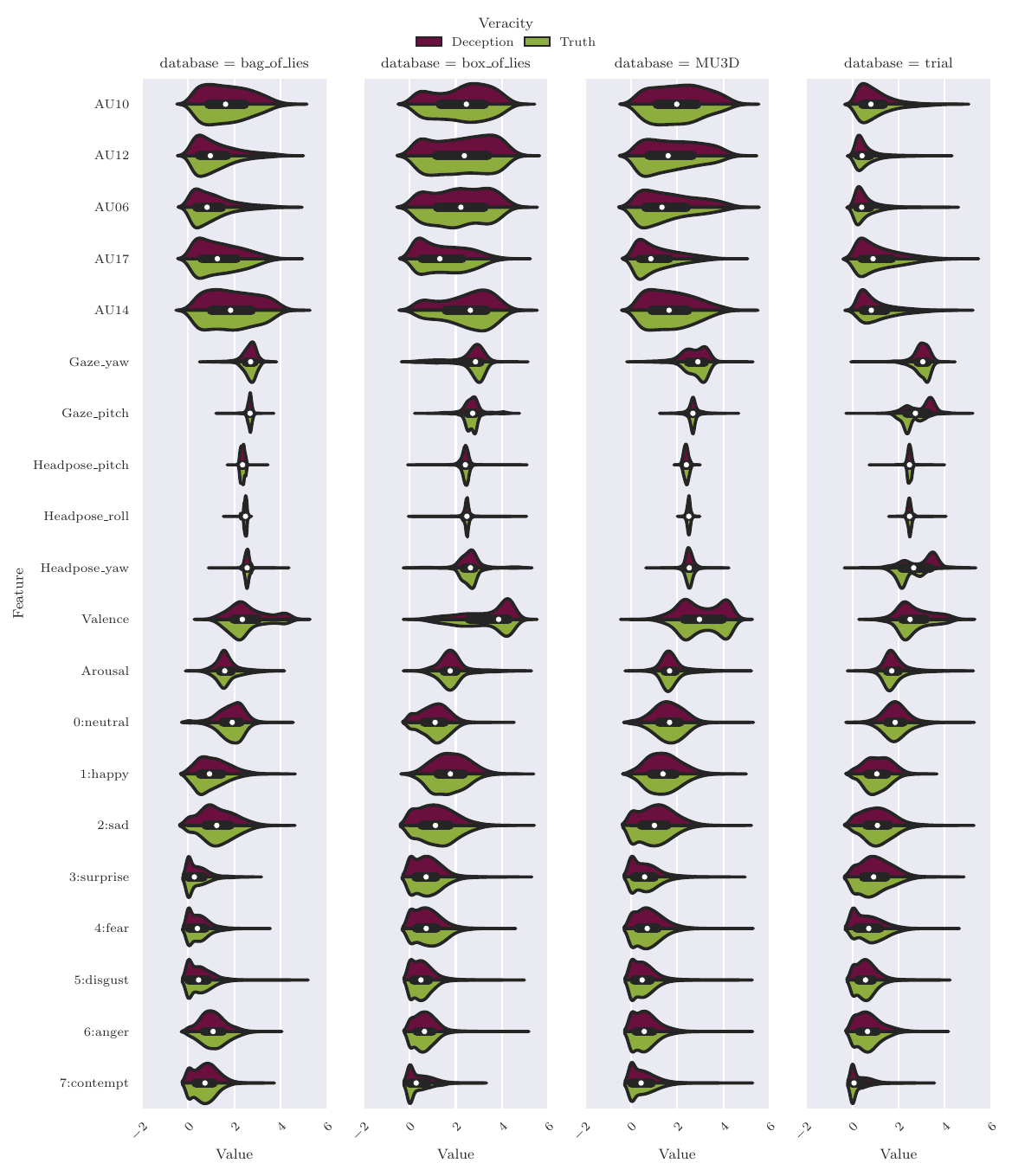}
	\caption[]{Violin plots show normalized distribution for all features.}	
	\label{fig:Features_violin_complete}
\end{figure*} 

\begin{figure*}[p]%
\centering
\includegraphics[]{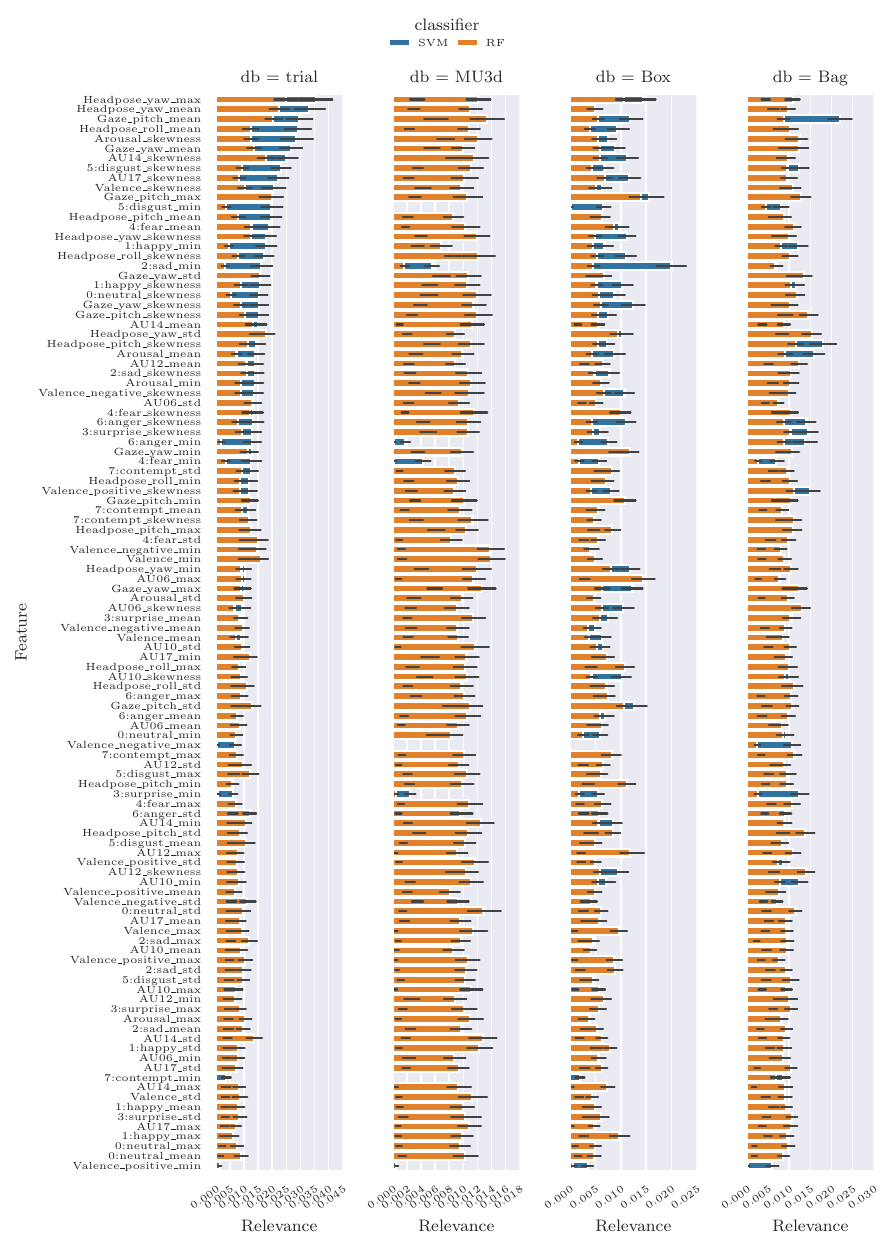}%
	\caption[]{
		 Feature relevance computed by feature permutation using traditional classifiers (SVM and Random Forest). Features are ordered according relevance for the trial database based on the SVM classifier. As one can see, the headpose and gaze features are expecially relevant in case of the trial database, but the feature relevance is very sensitive to the dataset. Furthermore, the relevance also depends on the classifier. 		
	}	
	\label{fig:FeatureRanking_all}
\end{figure*}

%MIGHT be deleted (not so important)
\begin{figure*}[tb!]%
\centering
\includegraphics[width=\textwidth]{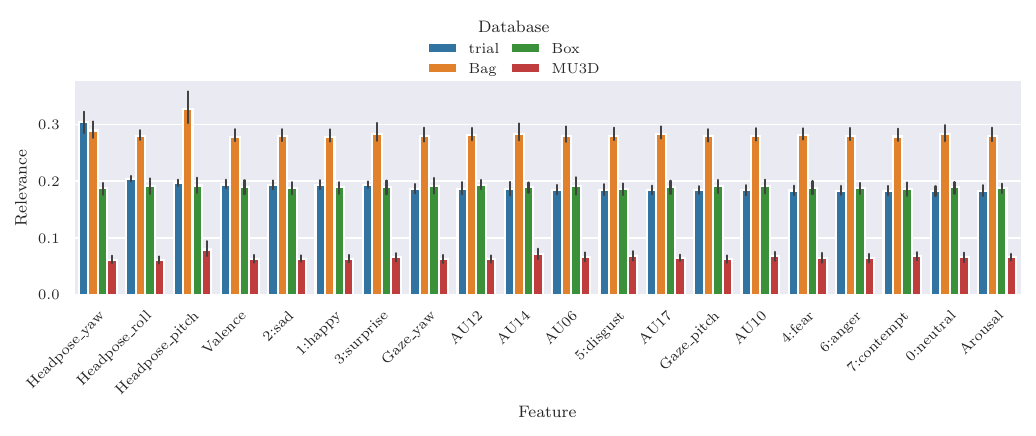}%
	\caption[]{
		 Feature relevance  for LSTM (ordered by Feature relevance for trial database, which is the only high-stake dataset). Feature relevance was computed by iteratively setting values for one feature to random values of its (Gaussian) distribution. AS one can see, only the Headpose (yaw) feature has a clearly higher relevance than the other features in the case of the trial database, while  features have a similar relevance in all other cases.		
	}	
	\label{fig:VAmap_affectnet}
\end{figure*}




\end{appendices}
\FloatBarrier


\bibliographystyle{sn-basic}% common bib file
%\bibliography{Ref}


\end{document}